%% file: main.tex
\documentclass{applemlr}
\usepackage{amsmath}
\usepackage{enumerate} 
\usepackage{algorithm}
\usepackage{algpseudocode}
\usepackage{amsfonts}
\usepackage{amsthm}
\usepackage{newtxtt}
\usepackage{cleveref}
\usepackage{diagbox} 
\usepackage{colortbl}
\usepackage{amssymb}
\usepackage{xspace}
\usepackage{wrapfig}
\usepackage{adjustbox}
\usepackage{tabularx}
\usepackage{booktabs}
\usepackage{mathtools}
\usepackage{tikz}
\usepackage{enumitem}
\usepackage{silence}
\usepackage{dsfont}
\usepackage[table]{xcolor}
\usepackage[dvipsnames]{xcolor}
\usepackage{multirow}
\usepackage{makecell}
\usepackage{xfakebold}
\usepackage[T1]{fontenc}
\input{math_commands}

\definecolor{textgray}{HTML}{6E6E73}
\usetikzlibrary{positioning, calc}
\usetikzlibrary{decorations.pathmorphing}

\makeatletter
\patchcmd{\wrong@fontshape}{\@gobbletwo}{}{}{}
\makeatother
\makeatletter
\AtBeginDocument{%
  \urlstyle{sf}
  
}
\makeatother

\numberwithin{equation}{section} 
\definecolor{light}{RGB}{125, 125, 125}
\crefname{tcb@cnt@pbox}{code}{code}
\Crefname{tcb@cnt@pbox}{Code}{Code}
\crefname{assumption}{assumption}{assumption}
\Crefname{assumption}{Assumption}{Assumptions}

\newtcolorbox[auto counter]{pbox}[2][]{
  colback=white,
  title=Code~\thetcbcounter: #2,
  #1,fonttitle=\sffamily,
  fontupper=\sffamily,
  arc=2pt,
  colframe=bgcolor,
  coltitle=fgcolor,
  colbacktitle=bgcolor,
  toptitle=0.25cm,
  bottomtitle=0.125cm
}

\makeatletter
\newcommand\applefootnote[1]{%
  \begingroup
  \renewcommand\thefootnote{}%
  \renewcommand\@makefntext[1]{\noindent##1}%
  \footnote{#1}%
  \addtocounter{footnote}{-1}%
  \endgroup
}
\makeatother

\definecolor{cverbbg}{gray}{0.90}

\newcommand{\framework}{MixAtlas\xspace}

\title{MixAtlas: Uncertainty-aware Data Mixture
Optimization for Multimodal LLM Midtraining}

\author{
Bingbing Wen$^\star$$^1$, Sirajul Salekin$^\circ$, Feiyang Kang$^\dagger$, Bill Howe$^\star$, Lucy Lu Wang$^\star$, Javier Movellan$^\circ$,Manjot Bilkhu$^\circ$ \\
}

\affiliation{$^\circ$Apple,  $^\star$ University of Washington, $^\dagger$ Virginia Tech, $^1$Work done during an internship at Apple}

\abstract{Domain reweighting can improve sample efficiency and downstream generalization, but data-mixture optimization for multimodal midtraining remains largely unexplored. Current multimodal training recipes tune mixtures along a single dimension, typically data format or task type. We introduce \framework, a method that produces benchmark-targeted data recipes that can be inspected, adapted, and transferred to new corpora. \framework decomposes the training corpus along two axes: \emph{image concepts} (10 visual-domain clusters discovered via CLIP embeddings) and \emph{task supervision} (5 objective types including captioning, OCR, grounding, detection, and VQA). Using small proxy models (Qwen2-0.5B) paired with a Gaussian-process surrogate and GP-UCB acquisition, \framework searches the resulting mixture space with the same proxy budget as regression-based baselines but finds better-performing mixtures. We evaluate on 10 benchmarks spanning visual understanding, document reasoning, and multimodal reasoning. On Qwen2-7B, optimized mixtures improve average performance by 8.5\%--17.6\% over the strongest baseline; on Qwen2.5-7B, gains are 1.0\%--3.3\%. Both settings reach baseline-equivalent training loss in up to 2$\times$ fewer steps. Recipes discovered on 0.5B proxies transfer to 7B-scale training across Qwen model families.
}

\date{\sffamily\today}

\begin{document}

\maketitle

\input{sections/1_introduction}
\input{sections/6_related_work}
\input{sections/2_method}
\input{sections/3_experiments}
\input{sections/4_results}

\input{sections/5_conclusion}


\applefootnote{ \textcolor{textgray}{\sffamily Apple and the Apple logo are trademarks of Apple Inc., registered in the U.S. and other countries and regions.}}

\clearpage
\newpage
\bibliographystyle{plainnat}
\bibliography{biblio}


\end{document}

%% file: math_commands.tex
\usepackage{amsmath,amsfonts,bm}

\def\eqref#1{equation~\ref{#1}}

\def\1{\bm{1}}

\DeclareMathAlphabet{\mathsfit}{\encodingdefault}{\sfdefault}{m}{sl}
\SetMathAlphabet{\mathsfit}{bold}{\encodingdefault}{\sfdefault}{bx}{n}



%% file: sections/1_introduction.tex
\section{Introduction} \label{section:introduction}

Multimodal large language models (MLLMs)~\citep{liu2023llava, chen2023sharegpt4v, deitke2025molmo, beyer2024paligemma} are increasingly used as the backbone for vision--language applications. A central but underexplored question in this setting is how to compose training data across heterogeneous visual concepts and multimodal objectives. The question is particularly relevant during \emph{midtraining}, when models are trained on high-resolution images and curated annotations to acquire a broad set of vision--language capabilities.

However, data mixture optimization for MLLMs is underexplored. Most MLLMs still rely on simple heuristics~\citep{shukor2025scaling, McKinzie2024MM1MAC,Baietal2024,Liu2025MidtrainingBPAU,roth2024a}, largely because systematic exploration is expensive: even a modest search over mixture weights can require dozens to hundreds of training runs, each with substantial compute cost, and the resulting mixtures are often difficult to interpret or transfer across model scales.

In this work, we introduce \framework, a framework for interpretable and compute-efficient multimodal mixture optimization.
The core idea is to convert an unstructured collection of midtraining datasets into an explicit, controllable data decomposition along two critical and interpretable axes. We optimize each axis independently; this decoupled design isolates the effect of each factor and keeps the search space low-dimensional.

Along the task supervision axis, we synthesize samples by user-defined objective type (e.g., detailed captioning, OCR), reflecting distinct supervision signals.
Along the image concept axis, we use large-scale clustering based on embeddings from the vision encoder to discover concepts automatically.
A \emph{mixture} in \framework is then a sampling distribution over each axis: at each step, training examples are drawn according to weights over task supervision or image concepts.
This decomposition makes mixture design interpretable: rather than treating each dataset as an opaque unit, \framework allows users to diagnose which visual domains or supervision signals are responsible for gains on specific downstream tasks, enabling targeted data collection and informed trade-off decisions when prioritizing capabilities.

To make mixture optimization practical under limited compute, \framework combines small proxy models with an uncertainty-aware policy.
We train lightweight proxy models on a small number of selected mixtures, then fit a Gaussian-process surrogate that predicts downstream performance for unseen mixtures while quantifying uncertainty.

Empirically, \framework yields significant improvements in both efficiency and accuracy.
Across a diverse benchmark suite, optimized mixtures consistently improve average performance by 8.5\%–17.6\% on Qwen2-7B and 1.0\%–3.3\% on Qwen2.5-7B across 10 diverse benchmarks compared with the strongest data mixture baselines (e.g., uniform sampling, RegMix~\citep{liu2024regmix}, Chameleon~\citep{xie2025chameleon}).
Moreover, the learned mixtures accelerate training, reaching a target loss in up to 2$\times$ fewer optimization steps.
Importantly, mixtures discovered on 0.5B proxy models transfer to larger-scale (7B) training, preserving both the convergence and accuracy benefits and enabling practical mixture optimization without expensive large-model search.

In summary, our contributions are:
\begin{itemize}[noitemsep, topsep=0pt, leftmargin=10pt]
\item \textbf{Interpretable benchmark-driven data recipes.} We formulate multimodal midtraining mixture optimization as deriving interpretable, benchmark-targeted recipes and introduce a two-axis data decomposition over \emph{task supervision} and \emph{image concepts}. This turns an unstructured corpus into a controllable mixture space that users can inspect, adapt, and extend to their own corpora and downstream targets.

\item \textbf{Uncertainty-aware proxy-based search for multimodal mixtures.} We propose a practical mixture optimization procedure that combines small proxy training with a GP surrogate to efficiently explore the two-axis data decomposition under limited compute. Using the same proxy budget, the GP-based search finds better-performing mixtures compared with other baselines.

\item \textbf{Systematic empirical study of multimodal midtraining mixture optimization.} We adapt the closest LLM-only data mixture baselines to the multimodal setting and benchmark them in this regime, showing that \framework delivers consistent gains in downstream performance, up to 2$\times$ faster convergence, and transferable recipes from 0.5B proxy models to 7B-scale models across two model families.
\end{itemize}

\begin{figure*}[t]
\centering
\includegraphics[width=\linewidth]{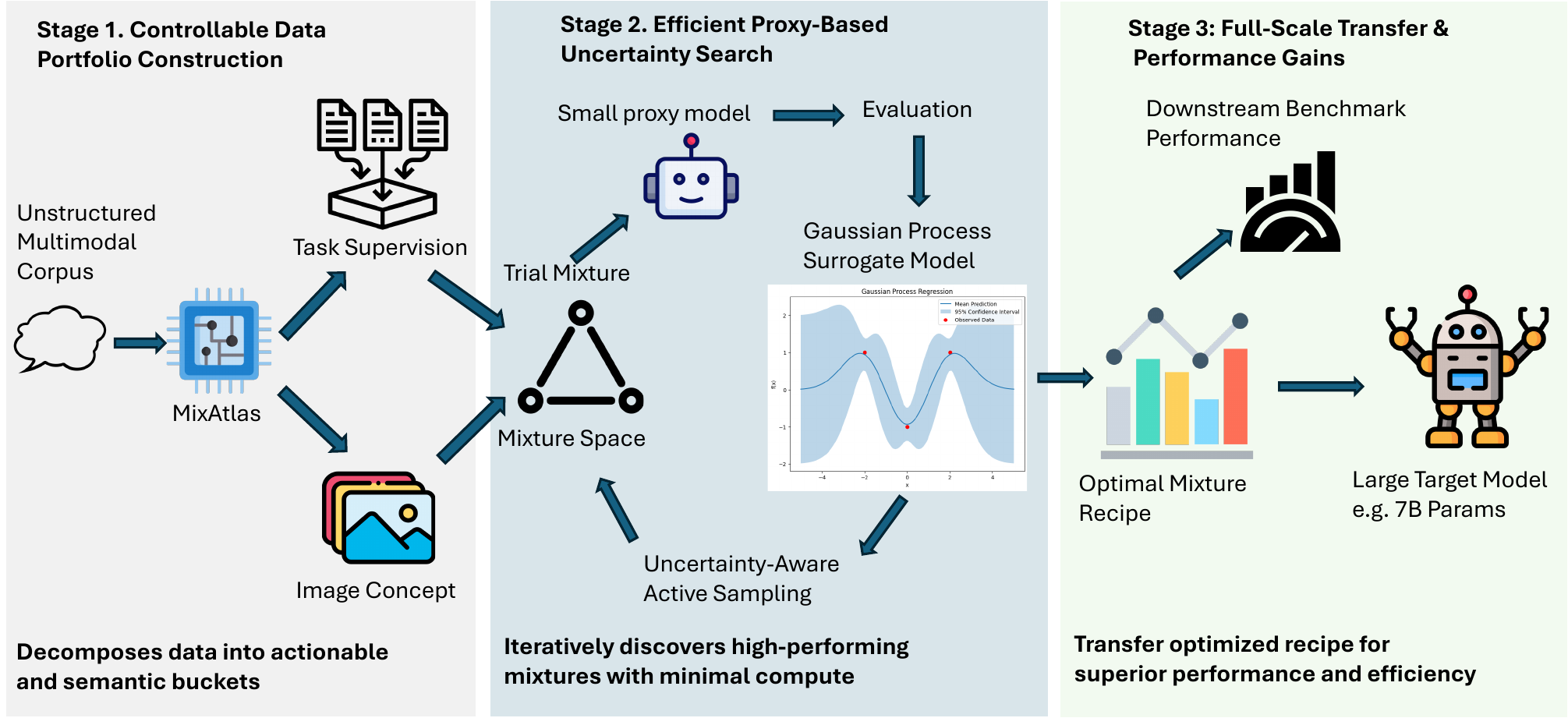}
\caption{Overview of \framework.
Stage 1: MixAtlas converts an unstructured multimodal corpus into a controllable data decomposition by decomposing examples along two interpretable axes: \emph{task supervision} and \emph{image concepts}. 
Stage 2: We efficiently explore the resulting mixture space using small proxy models and a Gaussian-process surrogate, guided by uncertainty-aware active sampling. 
Stage 3: The best mixture recipe transfers to full-scale training (e.g., 7B parameters), yielding faster convergence and improved downstream benchmark performance.}
\label{fig:framework}
\end{figure*}

%% file: sections/6_related_work.tex
\section{Related Work}
\noindent\textbf{Data Mixture Optimization in Pretraining/Midtraining.}
Data composition has long been recognized as a key driver of pretraining efficiency and downstream generalization, and recent work has moved from heuristic mixing toward \emph{principled} mixture optimization---primarily in language-only settings~\citep{kang2024autoscale}.
DoReMi~\citep{Xie2023DoReMi} formulates domain reweighting via distributionally robust optimization, learning mixture weights that improve worst-case domain performance.
RegMix~\citep{liu2024regmix} proposes proxy-based mixture search, showing that mixtures optimized on small models can transfer to larger scales.
Chameleon~\citep{xie2025chameleon} estimates domain importance using leverage scores in a learned embedding space, providing another mechanism for data selection and weighting.
Complementary analyses study how domain weights shape scaling behavior and performance trade-offs~\citep{shukor2025scaling}.

In multimodal pretraining and midtraining, mixture design is comparatively less explored and often conducted at a coarse granularity.
DataComp~\citep{gadre2023datacomp} performs large-scale studies of filtering and mixing for CLIP-style contrastive training, but focuses on image--text matching rather than multi-objective generative training.
Other work studies mixture heuristics over \emph{data formats} (e.g., caption vs.\ interleaved vs.\ text), finding fixed ratios that work well across several settings (e.g., the 5:5:1 recipe in MM1~\citep{McKinzie2024MM1MAC}), with synthetic data further improving certain regimes~\citep{Baietal2024,Liu2025MidtrainingBPAU, wen2023infovisdial, yang2025escaping, yao2025mmmg}.
Separately, \citet{roth2024a} examines the effect of \emph{data ordering} in continual pretraining.
Overall, existing multimodal recipes typically optimize mixtures from a \emph{single perspective} (format or task), rely on inherited ratios or expensive sweeps, and provide limited interpretability about which components drive which downstream gains.

Our work differs in three ways: (i) we focus on the \emph{midtraining} stage where multimodal capabilities are shaped by both visual content and supervision signals; (ii) we explicitly target \emph{interpretability} by decomposing the corpus into a controllable portfolio; and (iii) we perform \emph{benchmark-driven} mixture optimization in a structured space that captures interactions between visual concepts and training objectives.

\noindent\textbf{Domain Discovery and Characterization.}
A related line of work aims to define or discover ``domains'' in large datasets.
Prior work in visual domain adaptation studies distribution shift across styles and content~\citep{peng2019domain}, while foundational vision-language models demonstrate that semantic structure and visual concepts emerge in learned representations~\citep{radford2021learning}.
Nemotron-CLIMB~\citep{diao2025nemotronclimb} embeds and clusters large-scale data in a semantic space, then iteratively searches for mixtures using a proxy model and predictor.
Other efforts treat supervision type as the domain (e.g., PixMo~\citep{deitke2025molmo} and PaliGemma~\citep{beyer2024paligemma}), or organize web-scale corpora into topic/format taxonomies~\citep{wettig2025organize}.
While these approaches provide useful structure, they typically emphasize a \emph{single axis} (semantic clusters \emph{or} task type) and do not directly support fine-grained, jointly controllable mixtures that account for interactions between \emph{what} is seen (visual concepts) and \emph{how} it is supervised.

In contrast, our method proposes a two-axis decomposition (\emph{image concepts} $\times$ \emph{task supervision}) to obtain fine-grained multimodal mixture knobs, enabling both interpretable mixture control and attribution of downstream behavior to portfolio components.

\noindent\textbf{Efficient Hyperparameter Optimization.}
Searching over mixture weights is a challenging hyperparameter optimization problem due to the high-dimensional simplex structure and the cost of each training run.
Bayesian optimization~\citep{snoek2012practical} provides uncertainty-aware search but can struggle as mixture dimensionality grows and evaluations are expensive.
Recent work explores gradient-based data-mixture optimization~\citep{wang2023datamixture} and jointly learning mixture weights with model parameters~\citep{du2022mixturejoint}, trading off scalability, and compute cost.
Our approach is most closely related to proxy-based transfer methods such as RegMix~\citep{liu2024regmix} and $\mu$Transfer~\citep{yang2022mutransfer}, which show that hyperparameters optimized on smaller models can transfer to larger ones.

We extend this line to multimodal midtraining by combining proxy models with a Gaussian-process surrogate and uncertainty-aware active sampling, and by adapting mixture optimization to the multi-objective, cross-modal setting.



%% file: sections/2_method.tex
\section{Problem Formulation}
\label{sec:formulation}

Let $\widetilde{\mathcal{D}}=\{\widetilde{D}_1,\dots,\widetilde{D}_m\}$ denote the set of \emph{raw} midtraining datasets.
Each example is an image--text instance $(x^v, x^t)$.
Midtraining learns model parameters $\theta$ by minimizing the expected training loss under a sampling distribution
over data sources.


\paragraph{Mixture and training objective.}
A \emph{mixture} is a probability vector $h\in\Delta^{d-1}$ over portfolio elements:
$h_j\ge 0$ and $\sum_{j=1}^d h_j = 1$.
Given $h$, the midtraining objective is
\begin{equation}
\label{eq:train_obj}
\min_\theta \;\; \mathcal{L}(\theta; h)
= \mathbb{E}_{j\sim\text{Cat}(h)} \;\mathbb{E}_{(x,y)\sim P_j}\bigl[\ell(\theta; x, y)\bigr],
\end{equation}
where $\ell$ is the standard autoregressive token-level loss used by modern generative MLLMs.

\paragraph{Downstream selection objective.}
Let $\mathcal{B}$ be a suite of downstream benchmarks. Training with mixture $h$ yields a model $\theta(h)$ and a
benchmark score vector $F(h)=\{F_b(h)\}_{b\in\mathcal{B}}$.
MixAtlas supports two intuitive optimization modes:
\begin{itemize}[noitemsep, topsep=0pt, leftmargin=12pt]
\item \textbf{Generalist recipe:} maximize a weighted aggregate $\;J(h)=\sum_{b\in\mathcal{B}} w_b\,F_b(h)$
(with user-defined weights $w_b$ or uniform weights).
\item \textbf{Targeted recipe:} maximize a single benchmark $F_{b^\star}(h)$ for a specific capability (e.g., ChartQA).
\end{itemize}
In both cases, the goal is to find a mixture $h^\star$ that performs well under a fixed midtraining compute budget.

\section{Methodology}
\label{sec:method}
MixAtlas consists of three components as shown in Figure~\ref{fig:framework}: (i)Two-axis domain decomposition into task supervision and image concepts (\S\ref{sec:decomposition}),
yielding an interpretable data collection.(ii)Proxy-based, uncertainty-aware mixture optimization(\S\ref{sec:proxy_opt}) that searches the mixture
space using small proxy models and a probabilistic surrogate. (iii) Mixture transfer to full-scale midtraining(\S\ref{sec:transfer}), producing confidence-rated recipes.

\vspace{-2mm}

\subsection{Two-Axis Domain Decomposition}
\label{sec:decomposition}

We decompose the midtraining data along two independent axes (task supervision and image concepts) and optimize each axis independently: 
Given the decomposition, we form two portfolios:
(i) a task portfolio $\mathcal{P}_\text{task}=\{P_t: t\in\mathcal{T}\}$ and
(ii) a image concept portfolio $\mathcal{P}_\text{concept}=\{P_c: c\in\mathcal{C}\}$.
Each $P_t$ (resp. $P_c$) is the subset of examples with that task label (resp. concept cluster id).
A mixture $h$ then directly corresponds to interpretable sampling weights over tasks or concepts.

\subsubsection{Task Axis: Training task supervision}
\label{sec:task_axis}

We augment each image example by a task type $t\in\mathcal{T}$ and standardize all tasks into a unified instruction-following format: \texttt{<image> <instruction> → <response>}. We explain how we synthesize various task types based on the same image corpus as follows: 

\noindent\textbf{Detailed Captioning.}
We run an in-house model to generate a comprehensive natural-language description capturing objects, attributes, relations, and scene context. \textit{Template:} \texttt{<image> Describe this image in detail. → <caption>}

\noindent\textbf{OCR.} We run an off-the-shelf OCR engine on each image and concatenate the
extracted snippets in raster order. \textit{Template:} \texttt{<image> What text appears in this image? → <ocr\_tokens>}

\noindent\textbf{Grounded Captioning.}
We pair captions with object regions obtained from an in-house model, encoding each region as normalized coordinate tokens. \textit{Template:} \texttt{<image> Describe this image with grounded regions.
→ <caption> <ymin><xmin><ymax><xmax>}

\noindent\textbf{Detection.} We generate object labels using in-house model(Pix2Seq-style serialization of label–box pairs). \textit{Template:} \texttt{<image> Detect all objects.→ <label><ymin><xmin><ymax><xmax>}

\noindent\textbf{VQA.} We use the labels in LLaVA-Next midtraining corpus as-is for VQA. \textit{Template:} \texttt{<image> <question> → <answer>}

We consider the set of tasks above to construct a diverse set of training objectives and capabilities that are typically reflective of multimodal midtraining. However, MixAtlas is not limited to these task types and can be extended to support any task types that are of interest.

\subsubsection{Image Concept Axis: CLIP-Space Concept Clustering}
\label{sec:concept_axis}

To obtain an interpretable notion of ``visual domain'' beyond
dataset labels, we cluster image embeddings with $k$-means.

\noindent\textbf{CLIP embeddings.}
For each image $x^v$, we compute a vision embedding
$z^v=\phi(x^v)\in\mathbb{R}^d$ using a pretrained
CLIP~\citep{radford2021learning} vision encoder (e.g.,
ViT-L/14 at 336 resolution). We L2-normalize embeddings
prior to clustering, which improves stability for cosine
similarity.

\noindent\textbf{Scalable clustering.}
We fit $k$-means with $k=10$ on the embedding set (or on a
large random subset if the corpus is very large), then assign
every image to its nearest centroid. We choose $k=10$ as a
balance between granularity and interpretability.

\begin{wrapfigure}{r}{0.4\textwidth} %
\centering
\includegraphics[width=\linewidth]{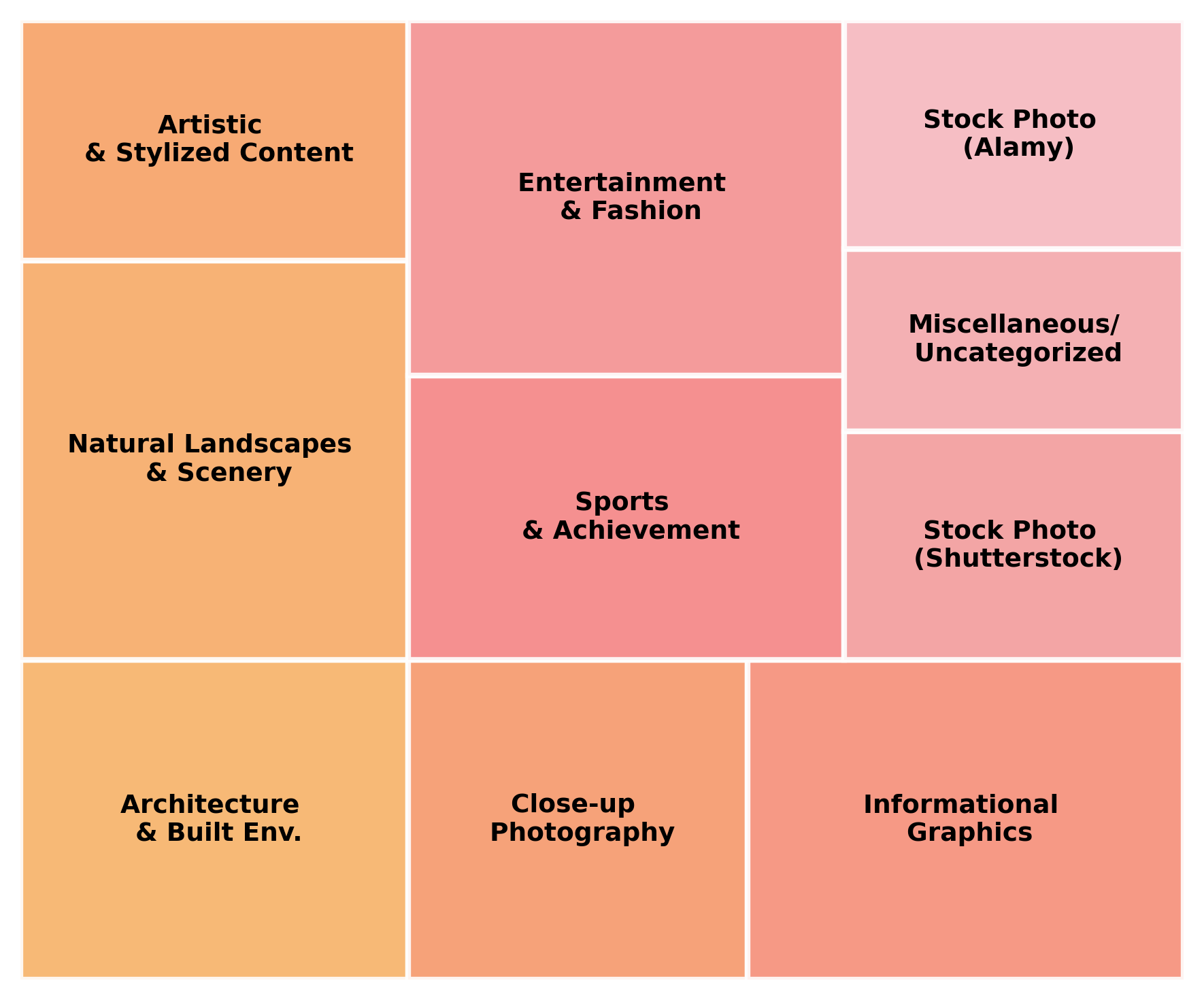}
\caption{Treemap of the ten automatically discovered image
  concept clusters, where each tile's area is proportional to
  the number of images assigned to that cluster. Cluster labels
  are assigned post-hoc by human annotators inspecting random
  samples.}
\label{fig:image_distribution}
\end{wrapfigure}

\noindent\textbf{Cluster naming and verification.}
The ten clusters are discovered automatically via $k$-means
and are not predefined by human taxonomy. To assign
human-readable labels, annotators inspect a random small sample of
images from each cluster and name the cluster according to its
dominant visual content. This post-hoc labeling yields the
following concept axes, whose relative sizes are shown in
Figure~\ref{fig:image_distribution}:
(1)~artistic \& stylized content,
(2)~entertainment \& fashion,
(3)~natural landscapes \& scenery,
(4)~sports \& achievement,
(5)~architecture \& built environments,
(6)~close-up photography,
(7)~informational graphics (charts, diagrams, infographics),
(8)~stock-photo(Alamy)
(9)~stock-photo(Shutterstock)
(10)~miscellaneous/uncategorized.


\subsection{Proxy-Based, Uncertainty-Aware Mixture Optimization}
\label{sec:proxy_opt}

Directly searching for $h^\star$ by repeatedly midtraining a large target model is prohibitively expensive.
MixAtlas therefore performs search using small proxy models and an uncertainty-aware surrogate. 
\vspace{-2mm}

\subsubsection{Candidate Pool Generation}
\label{sec:candidates}

We first generate a large pool of candidate mixtures $\mathcal{H}=\{h^{(1)},\dots,h^{(M)}\}\subset\Delta^{d-1}$ to ensure coverage
of the simplex. We combine:
\begin{itemize}[noitemsep, topsep=0pt, leftmargin=12pt]
\item Latin hypercube sampling on $\Delta^{d-1}$ for space-filling coverage, and
\item Dirichlet sampling $h\sim\text{Dir}(\alpha)$ with multiple concentration parameters $\alpha$ to include both
specialized mixtures (small $\alpha$) and near-uniform mixtures (large $\alpha$).
\end{itemize}
This pool acts as a discrete approximation to the continuous mixture space that we can score and search efficiently.

\subsubsection{Proxy Training and Benchmark Evaluation}
\label{sec:proxy_train}

For any candidate mixture $h$, we train a proxy model $M_h$ (e.g., 0.5B parameters) using the same training recipe as the target model
(architecture family, tokenizer, loss, and data formatting), but at reduced scale. All proxy runs use a fixed midtraining budget
(e.g., a fixed number of samples/steps), ensuring fair comparison across mixtures.

We then evaluate the proxy on the evaluation suite $\mathcal{V}$ aligned with the downstream target(s). Let
\begin{equation}
\label{eq:eval}
s(h) = \text{Eval}(M_h,\mathcal{V})
\end{equation}
denote the scalar objective used for search (e.g., a weighted aggregate of normalized benchmark scores, or a single benchmark score for a targeted recipe).

\subsubsection{Surrogate Model with Uncertainty}
\label{sec:surrogate}

We fit a probabilistic surrogate $g(h)$ to predict proxy performance from mixture weights, using the set of evaluated points
$\mathcal{S}=\{(h^{(i)}, s^{(i)})\}$ collected so far. We use a Gaussian process (GP) regressor, which yields a posterior predictive mean
$\mu(h)$ and standard deviation $\sigma(h)$:
\begin{equation}
g(h)\mid\mathcal{S}\;\Rightarrow\;\mu(h)\;\;,
\sigma(h)\;\;.
\end{equation}
Large $\sigma(h)$ indicates regions poorly supported by existing observations, which is exactly where additional proxy evaluations
are most informative.

\noindent\textbf{Interpretability via regression.}
In addition to the GP used for search, we optionally fit a simple second-order regression model on the same evaluated set,
\begin{equation}
\label{eq:quad_reg}
\hat{f}(h) = \beta_0 + \sum_{j} \beta_j h_j + \sum_{j<k} \beta_{jk} h_j h_k,
\end{equation}
where $\beta_j$ captures the marginal contribution of domain $j$ and $\beta_{jk}$ captures pairwise synergies/antagonisms.
This model supports the domain-sensitivity analyses and heatmaps reported in experiments.

\subsubsection{Uncertainty-Aware Active Sampling Policy}
\label{sec:policy}

\begin{wrapfigure}{r}{0.5\textwidth} 
    \vspace{-20pt} 
    \begin{minipage}{\linewidth}
        \hrule height 0.8pt 
        \vspace{3pt}
        \scriptsize
        \captionof{algorithm}{  \scriptsize{MixAtlas uncertainty-aware proxy search}}
        \label{alg:mixatlas}
        \hrule height 0.5pt 
        \begin{algorithmic}[1]
            \Require Candidate pool $\mathcal{H}$, evaluation budget $T$, initial seed size $T_0$, acquisition $a(\cdot)$
            \State Initialize $\mathcal{S}\leftarrow\emptyset$, sample $T_0$ seed mixtures $\{h^{(i)}\}$ from $\mathcal{H}$
            \For{$i=1$ to $T_0$}
                \State Train proxy $M_{h^{(i)}}$ with fixed budget; evaluate $s^{(i)}=\text{Eval}(M_{h^{(i)}},\mathcal{V})$
                \State $\mathcal{S}\leftarrow \mathcal{S}\cup\{(h^{(i)}, s^{(i)})\}$
            \EndFor
            \For{$t=T_0+1$ to $T$}
                \State Fit/update GP surrogate on $\mathcal{S}$; compute $(\mu(h),\sigma(h))$ for $h\in\mathcal{H}\setminus\mathcal{H}_\text{eval}$
                \State Select $h^{(t)}=\arg\max a(h)$ (e.g.,$\mu(h)+\kappa\sigma(h)$)
                \State Train proxy $M_{h^{(t)}}$; evaluate $s^{(t)}$
                \State $\mathcal{S}\leftarrow \mathcal{S}\cup\{(h^{(t)}, s^{(t)})\}$
            \EndFor
            \State \Return recommended recipe $\hat{h}=\arg\max_{(h,s)\in\mathcal{S}} s$ 
        \end{algorithmic}
        \vspace{2pt}
        \hrule height 0.8pt 
    \end{minipage}
    \vspace{-10pt} 
\end{wrapfigure}

Given the candidate pool $\mathcal{H}$ and current GP posterior, MixAtlas selects which mixtures to evaluate next under a limited
proxy budget.


\noindent\textbf{Exploration--exploitation via GP-UCB.}
To prioritize mixtures that are both promising and uncertain, we use an optimistic acquisition:
\begin{equation}
a_{\text{UCB}}(h) = \mu(h) + \kappa\,\sigma(h),
\end{equation}
where $\kappa$ controls exploration.

\vspace{-2mm}
\subsection{Mixture Transfer to Full-Scale Midtraining}
\label{sec:transfer}

Finally, we transfer the best proxy-discovered recipe $\hat{h}$ to the full-scale target model. The key assumption is that the \emph{relative ranking} of mixtures is preserved across model scales---that is, a mixture that outperforms alternatives at 0.5B parameters will also outperform them at 7B. This assumption is grounded in a growing body of evidence~\citep{liu2024regmix, kang2024autoscale,diao2025nemotronclimb} that data-level choices transfer more reliably across scales than optimizer hyperparameters. The intuition is that mixture weights control the \emph{distribution of supervision signals} seen during training, which shapes what the model learns to attend to. We empirically validate this transfer assumption in Section~\ref{main-results}, showing that recipes discovered on Qwen2-0.5B proxies yield consistent gains when applied to both Qwen2-7B and Qwen2.5-7B, including across model families (Qwen2 $\to$ Qwen2.5), with the largest improvements on benchmarks compared with proxy-based mixture methods.
 
\vspace{-3mm}

%% file: sections/3_experiments.tex
\section{Experiment Setup}
\noindent\textbf{Midtraining corpus.}
We construct a multi-task midtraining corpus by aggregating open-source multimodal datasets.
In our current setup, caption-style image--text supervision is primarily sourced from Conceptual Captions (CC3M~\citep{sharma2018conceptual} and CC12M~\citep{changpinyo2021cc12m}). We build on top of the mid-training dataset proposed by LLaVA-NeXT \citep{liu2023llava} and use the same image corpus but generate labels for other tasks (captioning, OCR, grounded captions, detection) using a combination of open-source and in-house models.
The original labels in the dataset are used for the VQA task as-is. For all other tasks, we generate labels for the entire dataset and use them to create different types of task supervision.
All datasets are converted into a unified instruction-following format (image + textual prompt $\rightarrow$ text tokens),
so that the same autoregressive loss can be used across tasks (see \S\ref{sec:method}).
We sample a fixed training budget per run (1 epoch, $\sim$4M samples; see below) so that comparisons differ \emph{only}
in the sampling distribution (mixture).

\noindent\textbf{Separate mixture optimization along task and concept axes.}
We study mixture optimization along two interpretable axes: (i) a \emph{task} portfolio with $|\mathcal{T}|=5$ supervision types (Dense captioning, OCR, Grounded captioning, Detection, VQA), and (ii) an \emph{image-concept} portfolio with $|\mathcal{C}|=10$ semantic clusters discovered by running $k$-means on CLIP image embeddings (\S\ref{sec:concept_axis}).  We empirically set k=5, 10, 20 and found that k=10 balanced granularity and search tractability. For concept clustering, we embed images using \texttt{openai/clip-vit-large-patch14-336} and L2-normalize embeddings before clustering. For either axis, a mixture is a probability vector over the corresponding portfolio, and we sample training examples according to these mixture weights.

Although the full data distribution can be viewed as a joint mixture over \emph{both} task types and image concepts, directly optimizing this joint space would require  substantial compute. We therefore decouple the axes and optimize them separately for controlled and interpretable comparisons: (i) a \emph{task recipe} that optimizes task weights while sampling concepts uniformly, and (ii) a \emph{concept recipe} that optimizes concept weights while sampling tasks uniformly. This design isolates the effect of each factor, enables axis-specific sensitivity analyses, and keeps proxy search tractable; we leave joint optimization over both axes to future work.

\noindent\textbf{Single-task vs. multi-task setup.}
For the single-task midtraining experiments, we train using only the original VQA task type in the LLaVA-Next datasets, while keeping the total training budget fixed to 1 epoch ($\sim$4M samples).
To ensure a fair comparison, we match the total number of training samples across single-task and multi-task runs. In the single-task case, all samples are drawn from that task domain; in the multi-task case, samples are drawn uniformly across the five task types (unless otherwise specified).
Therefore, differences between single-task and multi-task results reflect supervision diversity rather than total training volume.

\noindent\textbf{Models.}
We evaluate MixAtlas on the LLaVA-Next~\citep{liu2023llava} training pipeline with two 7B-scale language backbones:
Qwen2-7B and Qwen2.5-7B ~\citep{yang2025qwen3}.
For the vision encoder, we use the same CLIP model (\texttt{openai/clip-vit-large-patch14-336}) as in LLaVA-Next,
and we keep the overall architecture and training recipe fixed across mixtures.
To make mixture optimization compute-efficient, we use a smaller proxy model ( Qwen2-0.5B) to predict optimal mixtures and apply them to train bigger models (Qwen2-7B).

\noindent\textbf{Training setup.}
We follow the standard LLaVA-Next midtraining recipe.
Across all compared methods, we keep \emph{everything} fixed except the mixture weights:
model architecture, tokenizer, image resolution, maximum sequence length, optimizer/schedule configuration, and total training budget.
All full-scale midtraining runs are trained for 1 epoch ($\sim$4M samples) using $64\times$ H100 GPUs.

\input{table/combine_Qwen2_7b}
\input{table/combine_qwen2.5_7b}

\noindent\textbf{Evaluation suite and metrics.} We evaluate models on a broad suite spanning four categories: (i)Visual understanding: AI2D~\citep{Kembhavi2016ADI}, GQA~\citep{ainslie2023gqa}, ScienceQA~\citep{lu2022learn}. (ii) Document \& text: DocVQA~\citep{mathew2021docvqa}, TextVQA~\citep{singh2019towards}, ChartQA~\citep{masry-etal-2022-chartqa}. (iii) Multimodal reasoning: MMMU~\citep{yue2023mmmu}, MMBench~\citep{liu2024mmbench}, MM-Vet~\citep{yu2024mm}.
(iv) Mathematical reasoning: MathVista~\citep{lu2024mathvista}.

We use the standard metrics for each benchmark (e.g., accuracy for most VQA-style tasks, ANLS for DocVQA, and GPT-based evaluation where applicable),
and report all metrics on a 0--100 scale for readability. In this paper, we focus on the generalist recipe experiments. We use a uniform-weighted average of normalized benchmark scores as the scalar objective $s(h)$.

\noindent\textbf{Baselines.} To the best of our knowledge, no prior work addresses proxy-based mixture optimization for multimodal midtraining; existing methods~\citep{liu2024regmix, xie2025chameleon} target language-only pretraining. We adapt the most relevant of these to our multimodal pipeline, matching candidate pools, proxy budgets, and evaluation suites for fair comparison:
\begin{itemize}[noitemsep, topsep=0pt, leftmargin=12pt]
    \item \textbf{Uniform}: equal sampling probability over all portfolio elements (uniform over 5 task types for task-axis experiments; uniform over 10 image concepts for concept-axis experiments). This serves as the standard no-optimization reference.
    \item \textbf{Chameleon}~\citep{xie2025chameleon}: estimates domain importance via inverse scores computed in a learned image embedding space, then reweights sampling accordingly. Since leverage scores are defined over image representations, this baseline applies only to the concept axis; it has no natural analogue for the task axis where domains differ in supervision type rather than visual content.
    \item \textbf{RegMix}~\citep{liu2024regmix}: fits a LightGBM regression model that maps mixture weights to the average benchmark score, splits all proxy evaluations into training data and test data, then selects the mixture with the highest predicted score from the candidate pool. We adapt RegMix to our multimodal setting by replacing its original text domain  with our task and concept axis.
    \item \textbf{MixAtlas (ours)}: fits a Gaussian-process surrogate over the same mixture-to-benchmark-score mapping and selects candidates via GP-UCB, which balances exploitation of high-predicted regions with exploration of high-uncertainty regions. To ensure a fair comparison, MixAtlas and RegMix share the same candidate pool and the same proxy evaluation budget (50 runs for the task axis, 200 for the concept axis); the only difference is how each method selects which candidates to evaluate and how it models the performance landscape.
\end{itemize}

%% file: table/combine_Qwen2_7b.tex
\begin{table*}[t]
\centering
\setlength{\tabcolsep}{3pt}
\scriptsize
\caption{Comparison of Mid-training Strategies on Qwen2-7B on both Task Axis and Image Concept Axis. Gain is computed relative to the strongest baseline for each axis. Improvements (positive gains) and best performing methods are in bold.}
\resizebox{\textwidth}{!}{%
\begin{tabular}{l c c c >{\columncolor[HTML]{F2F2F2}}c c c c c c >{\columncolor[HTML]{F2F2F2}}c}
\toprule
\scriptsize
 & \multicolumn{4}{c}{\textbf{Task Axis}} & & \multicolumn{5}{c}{\textbf{Image Concept Axis}} \\
\cmidrule(lr){2-5} \cmidrule(lr){7-11}
\textbf{Benchmark} & \textbf{Unif} & \textbf{Reg} & \textbf{\framework} & \textbf{Gain} & &
\textbf{Unif} & \textbf{Reg} & \textbf{Cham} & \textbf{\framework} & \textbf{Gain} \\
\midrule
\multicolumn{11}{l}{\textit{Visual Understanding}} \\
AI2D      & 58.16 & 58.10 & \textbf{59.84} & \textbf{+2.9\%}  & & 54.30 & 53.00 & 56.50 & \textbf{57.20} & \textbf{+1.2\%} \\
GQA       & 18.95 & 20.90 & \textbf{30.65} & \textbf{+46.6\%} & & 25.14 & 23.35 & 23.90 & \textbf{36.48} & \textbf{+45.1\%} \\
ScienceQA & 72.53 & 72.30 & \textbf{74.81} & \textbf{+3.1\%}  & & 71.60 & \textbf{72.90} & 72.30 & 71.50 & -1.9\% \\
\midrule
\multicolumn{11}{l}{\textit{Document and Text}} \\
DocVQA    & 60.17 & 58.30 & \textbf{62.98} & \textbf{+4.7\%}  & & 3.50  & 15.80 & 3.00  & \textbf{46.90} & \textbf{+196.8\%} \\
TextVQA   & 50.03 & 48.10 & \textbf{54.51} & \textbf{+9.0\%}  & & 29.90 & 30.00 & 25.60 & \textbf{50.90} & \textbf{+69.7\%} \\
ChartQA   & 43.44 & 46.00 & \textbf{50.16} & \textbf{+9.0\%}  & & 33.30 & 38.20 & 34.60 & \textbf{41.80} & \textbf{+9.4\%} \\
\midrule
\multicolumn{11}{l}{\textit{Multimodal Reasoning}} \\
MMBench-EN & 63.40 & 63.50 & \textbf{67.35} & \textbf{+6.1\%}  & & 53.78 & \textbf{68.39} & 59.54 & 61.89 & -9.5\% \\
MMMU Val   & 39.78 & 39.80 & \textbf{41.67} & \textbf{+4.7\%}  & & 37.40 & 40.00 & 38.10 & \textbf{40.30} & \textbf{+0.8\%} \\
MM-Vet     & \textbf{28.12} & 26.10 & 27.75 & -1.3\%  & & 27.39 & 26.15 & 23.21 & \textbf{30.37} & \textbf{+10.9\%} \\
\midrule
\multicolumn{11}{l}{\textit{Mathematical}} \\
MathVista & 32.60 & 31.80 & \textbf{37.10} & \textbf{+13.8\%} & & 35.59 & 33.40 & \textbf{36.20} & 34.50 & -4.7\% \\
\midrule
\textbf{AVERAGE} & 46.72 & 46.50 & \textbf{50.68} & \textbf{+8.5\%} & & 37.19 & 40.12 & 37.30 & \textbf{47.18} & \textbf{+17.6\%} \\
\bottomrule
\end{tabular}
}
\label{tab:qwen2_7b_task_vs_image_concept}
\end{table*}

%% file: table/combine_qwen2.5_7b.tex
\begin{table*}[t]
\centering
\scriptsize
\setlength{\tabcolsep}{3pt}
\caption{Comparison of Mid-training Strategies on Qwen2.5-7B on Task Axis vs. Image Concept Axis. Gain is computed relative to the strongest baseline for each axis. Improvements (positive gains) and best performing methods are in bold.}
\resizebox{\textwidth}{!}{%
\begin{tabular}{l c c c >{\columncolor[HTML]{F2F2F2}}c c c c c c >{\columncolor[HTML]{F2F2F2}}c}
\toprule
 & \multicolumn{4}{c}{\textbf{Task Axis}} & & \multicolumn{5}{c}{\textbf{Image Concept Axis}} \\
\cmidrule(lr){2-5} \cmidrule(lr){7-11}
\textbf{Benchmark} & \textbf{Unif} & \textbf{Reg} & \textbf{\framework} & \textbf{Gain} & &
\textbf{Unif} & \textbf{Reg} & \textbf{Cham} & \textbf{\framework} & \textbf{Gain} \\
\midrule
\multicolumn{11}{l}{\textit{Visual Understanding}} \\
AI2D      & 58.65 & \textbf{60.80} & 59.97 & -1.4\%  & & 55.31 & \textbf{62.21} & 58.16 & 57.19 & -8.1\% \\
GQA       & 32.61 & 39.00 & \textbf{45.05} & \textbf{+15.5\%} & & 25.14 & \textbf{44.27} & 23.90 & 36.48 & -17.6\% \\
ScienceQA & \textbf{74.62} & 73.80 & 73.18 & -1.9\%  & & \textbf{74.07} & 73.53 & 73.33 & 71.49 & -3.5\% \\
\midrule
\multicolumn{11}{l}{\textit{Document and Text}} \\
DocVQA  & 56.03 & 66.40 & \textbf{74.31} & \textbf{+11.9\%} & & 22.68 & 24.05 & 25.86 & \textbf{46.83} & \textbf{+81.1\%} \\
TextVQA & 53.24 & 56.70 & \textbf{64.24} & \textbf{+13.3\%} & & 33.10 & 48.76 & 40.60 & \textbf{50.92} & \textbf{+4.4\%} \\
ChartQA & 52.84 & 53.50 & \textbf{53.96} & \textbf{+0.9\%}  & & 38.40 & 41.16 & 39.08 & \textbf{41.88} & \textbf{+1.7\%} \\
\midrule
\multicolumn{11}{l}{\textit{Multimodal Reasoning}} \\
MMBench-EN & 60.40 & \textbf{72.80} & 61.68 & -15.3\% & & 63.23 & \textbf{65.89} & 62.97 & 60.82 & -7.7\% \\
MMMU Val   & 41.11 & 42.00 & \textbf{42.67} & \textbf{+1.6\%}  & & 38.67 & \textbf{42.56} & 39.00 & 40.33 & -5.2\% \\
MM-Vet     & 31.38 & 22.60 & \textbf{31.51} & \textbf{+0.4\%}  & & 25.60 & 27.16 & 27.89 & \textbf{31.42} & \textbf{+12.7\%} \\
\midrule
\multicolumn{11}{l}{\textit{Mathematical}} \\
MathVista & 37.50 & \textbf{37.80} & 36.10 & -4.5\%  & & 33.90 & \textbf{38.00} & 34.90 & 34.70 & -8.7\% \\
\midrule
\textbf{AVERAGE} & 49.84 & 52.54 & \textbf{54.27} & \textbf{+3.3\%} & & 41.01 & 46.76 & 42.57 & \textbf{47.21} & \textbf{+1.0\%} \\
\bottomrule
\end{tabular}
}
\label{tab:qwen2_5_7b_task_vs_image_concept}
\end{table*}

%% file: sections/4_results.tex
\section{Results}
\label{main-results}

\framework consistently improves both effectiveness and efficiency across our evaluation suite.
Across 7B-scale midtraining runs, MixAtlas-learned mixtures (i) improve average benchmark performance over uniform and widely-used data mixture optimization baselines and
(ii) improve sample efficiency by reaching the same training loss in much fewer optimization steps.
Gains are largest on benchmarks whose requirements align strongly with specific visual concepts, while broad-coverage
benchmarks tend to benefit from more diversified concept mixtures.

\paragraph{\framework recipes outperform strong baselines at 7B scale}
We evaluate MixAtlas along each axis separately to isolate the effect of each axis, producing two recipes: a \emph{task recipe} (task-supervision weights) and a
\emph{concept recipe} (image-concept weights). 
The task-recipe experiments optimize supervision-type weights while fixing concept weights to uniform;
the concept-recipe experiments optimize concept weights while fixing task weights to uniform.
The evaluation suite is identical in both cases (same benchmarks and metrics).
The difference between the result tables lies solely in which mixture axis was optimized.
\begin{itemize}[noitemsep, topsep=0pt, leftmargin=10pt]
\item \textbf{Task recipe} Table~\ref{tab:qwen2_7b_task_vs_image_concept} shows that the \textsc{MixAtlas} task-optimal mixture improves the overall average from 46.72/46.50 (Uniform/RegMix) to 50.68 (+8.5\% relative gain over the stronger baseline) and outperforms the stronger of Uniform and RegMix on 9 of 10 benchmarks. The largest improvements occur on GQA (30.65 vs.\ 20.90, +46.6\%), MathVista (37.10 vs.\ 32.60, +13.8\%), and ChartQA/TextVQA (both +9.0\% relative). The only degradation is on MM-Vet ($-$1.3\%), highlighting that task reweighting can introduce trade-offs depending on the target. Crucially, both \textsc{MixAtlas} and RegMix use the same budget of \textbf{50 proxy runs} on the 5-dimensional task simplex, yet RegMix's regression-based selection underperforms \textsc{MixAtlas} on average by a clear margin (46.50 vs.\ 50.68). This gap highlights the value of uncertainty-aware search: rather than fitting a fixed regression model to all 50 observations and selecting the predicted optimum, \textsc{MixAtlas}'s GP-UCB policy actively steers proxy evaluations toward the most promising and uncertain regions of the simplex, extracting more information from the same compute budget.

\item \textbf{Concept recipe} Table~\ref{tab:qwen2_7b_task_vs_image_concept} shows that the \textsc{MixAtlas} concept-optimal mixture achieves the best overall average (47.18) compared to Uniform (37.19), RegMix (40.12), and Chameleon (37.30), a +17.6\% relative gain over the strongest baseline. The largest gains are on document/text benchmarks: DocVQA (46.90 vs.\ 15.80, +196.8\%), TextVQA (50.90 vs.\ 30.00, +69.7\%), and ChartQA (41.80 vs.\ 38.20, +9.4\%), while also improving MM-Vet (30.37 vs.\ 27.39, +10.9\%). The advantage of GP-UCB over regression-based search is even more pronounced on this harder problem: both methods draw from the same pool of \textbf{200 proxy runs} on the 10-dimensional concept simplex, yet RegMix's second-order regression model cannot adequately capture the performance landscape in this higher-dimensional space, leaving a 7-point gap to \textsc{MixAtlas} (40.12 vs.\ 47.18). The GP surrogate's ability to model non-linear interactions and focus exploration on high-uncertainty, high-reward regions becomes increasingly important as the dimensionality of the mixture space grows. Together, these results show that controlling both supervision composition and concept composition is important; mixtures that rely on fixed heuristics or lower-capacity surrogate models cannot match the task- and domain-aware optima identified by \textsc{MixAtlas}.
 
\end{itemize}

\paragraph{Recipes transfer across model families and scales}
A key advantage of MixAtlas is that recipes can be discovered using smaller proxy models and then transferred to much larger target models.
Table~\ref{tab:qwen2_5_7b_task_vs_image_concept} shows that transferring the learned recipe from Qwen2-0.5B proxies to Qwen2.5-7B yields clear gains
on multiple benchmarks, especially in document/text and fine-grained understanding: DocVQA improves from 56.03 (Uniform) to 74.31 (Optimal),
TextVQA from 53.24 to 64.24, and GQA from 32.61 to 45.05. We also observe trade-offs: MMBench-EN decreases relative to the
strongest baseline (72.80 to 61.68), and MathVista variants see modest drops. These patterns are consistent with the sensitivity maps above:
mixtures that prioritize text/document concepts can trade off broad-coverage benchmarks. We note that on the concept axis with the stronger base model (Qwen2.5-7B), gains are concentrated in document/text tasks (DocVQA +81.1\%, TextVQA +4.4\%, ChartQA +1.7\%) while visual understanding and reasoning benchmarks decline, narrowing the overall advantage over RegMix to +1.0\%. This suggests that as the base model improves, the marginal benefit of concept reweighting becomes more task-specific, reinforcing the value of MixAtlas as a target-aware recipe tool rather than a uniform-gain method.

\begin{figure*}[t]
    \centering
    \begin{minipage}[t]{0.48\textwidth}
        \centering
        \includegraphics[width=\linewidth]{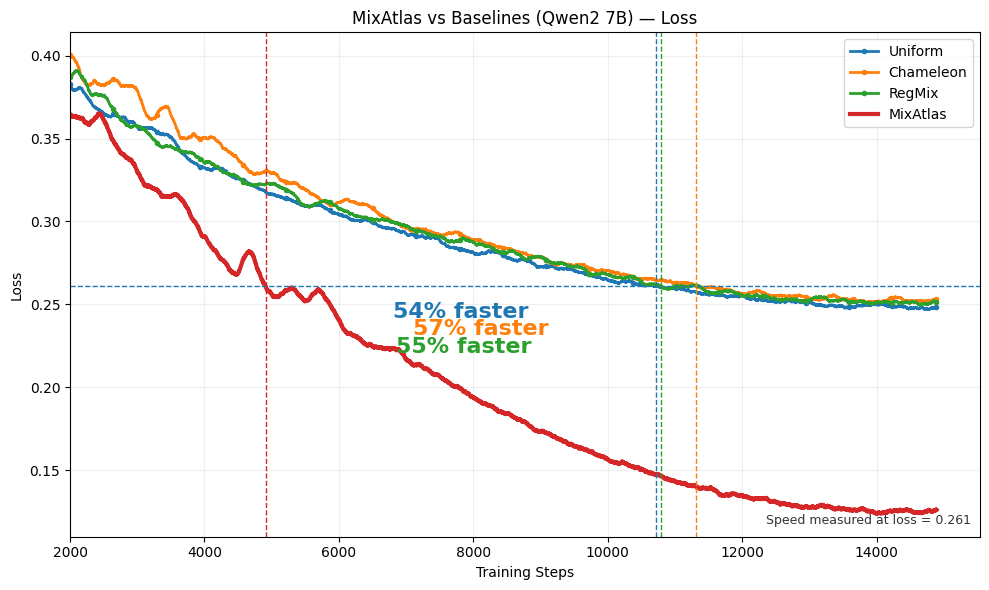}
        \caption{MixAtlas convergence efficiency in Qwen2 7B. It matches baseline final losses at least 54\% fewer steps.}
        \label{fig:final_loss_qwen2}
    \end{minipage}
    \hfill 
    \begin{minipage}[t]{0.48\textwidth}
        \centering
        \includegraphics[width=\linewidth]{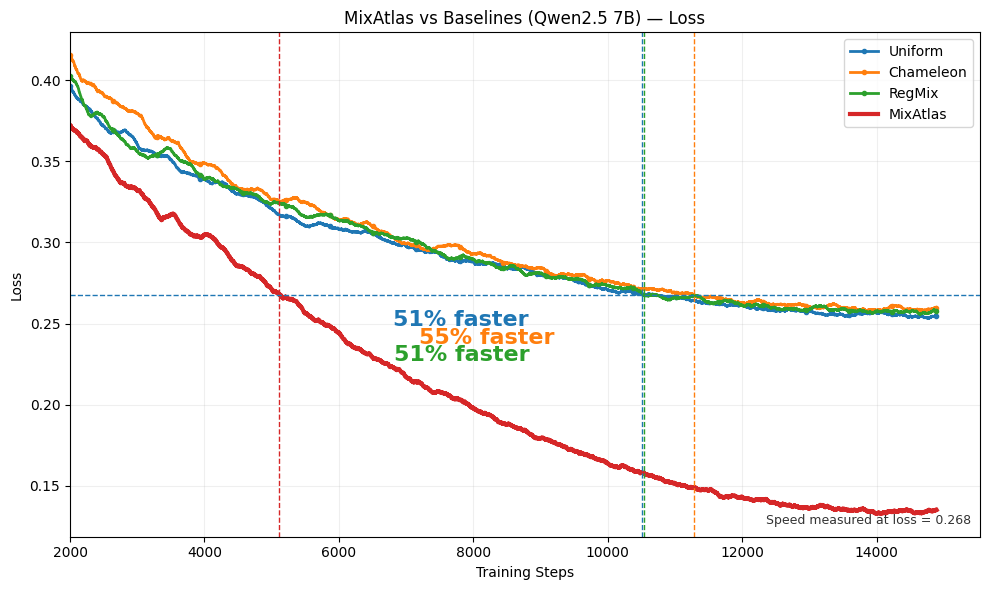}
        \caption{MixAtlas convergence efficiency in Qwen2.5 7B. It matches baseline final losses at least 51\% fewer steps.}
        \label{fig:final_loss_qwen2.5}
    \end{minipage}
\end{figure*}

\begin{wrapfigure}{r}{0.45\textwidth} %
  \centering
  \includegraphics[width=\linewidth]{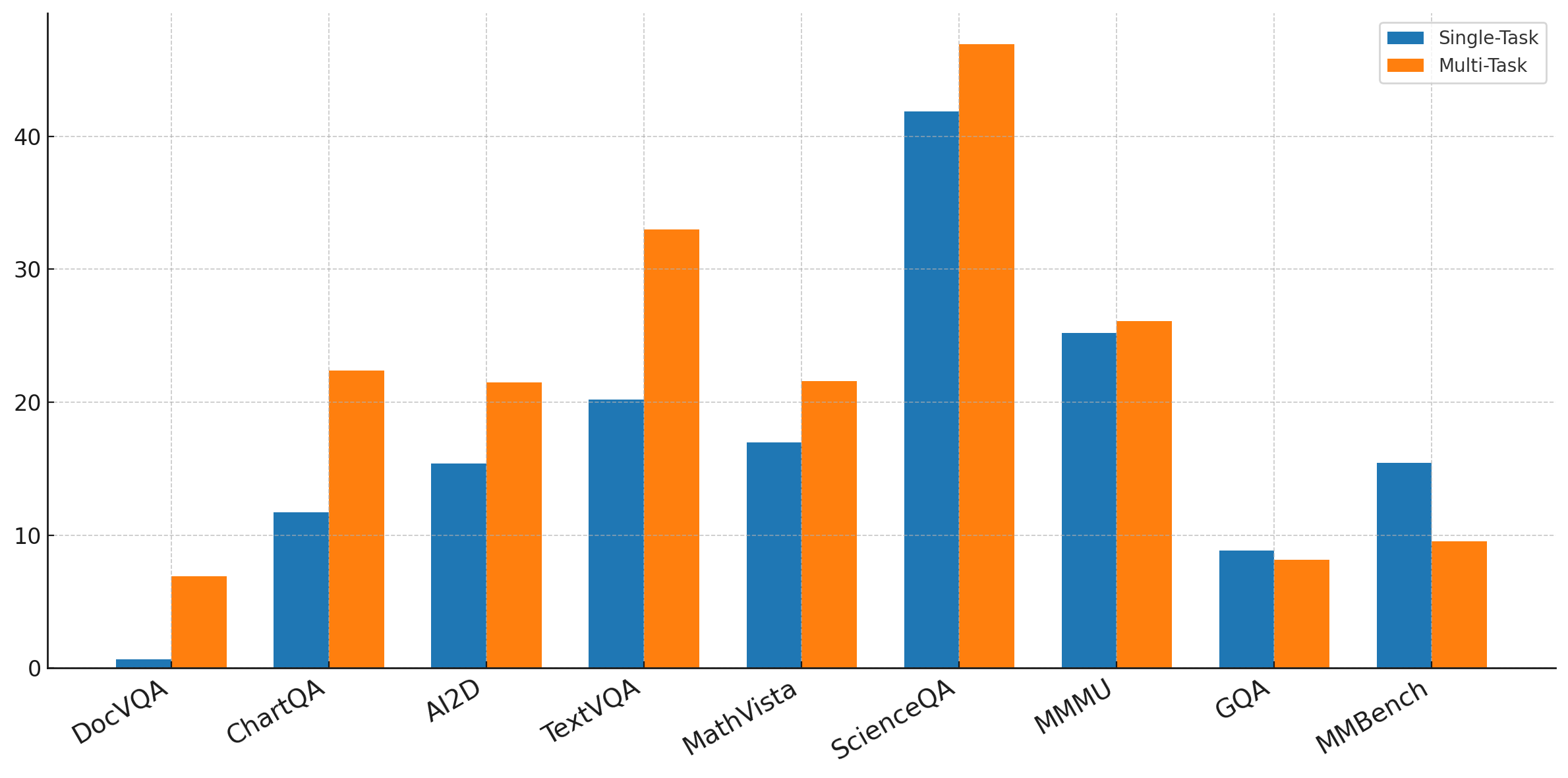}
  \caption{Single-task and multi-task midtraining strategies across benchmark tasks, revealing the benefits of incorporating diverse supervision signals during midtraining.}
  \label{fig:single_multi_task}
\end{wrapfigure}

\paragraph{\framework significantly improves training efficiency by reaching the same loss with at least 51\% fewer steps.}
MixAtlas does not only improve final accuracy; it also accelerates optimization.
On the target model (LLaVA-Next Qwen2-7B) in Figure~\ref{fig:final_loss_qwen2} , the MixAtlas-optimal mixture reaches the same loss in 54\% fewer steps than Uniform,  57\% fewer
steps than Chameleon and 55\% fewer steps than RegMix. Figure~\ref{fig:final_loss_qwen2.5} shows that at a matched loss level (dashed reference),
the optimized mixture reaches the target in roughly 4k steps, while Uniform and Chameleon require close to 11k steps.
The loss gap widens as training progresses, indicating that mixture optimization improves both early learning speed and overall optimization efficiency.

\section{Analysis}
\label{analysis}

\paragraph{Multi-task mixtures surpass single-task midtraining.} 
Beyond optimizing weights, we examine whether supervision diversity itself provides gains.
Specifically, we compare single-task midtraining (same total data budget, drawn from only one task type) against uniform multi-task midtraining (same total budget, distributed evenly across tasks).

As shown in Figure~\ref{fig:single_multi_task}, multi-task midtraining (captioning + OCR + grounding + VQA + Detection) outperforms single-task training
on 8 of 9 benchmarks, with especially large gains on text-heavy benchmarks: ChartQA increases from 12.0 to 22.4 (+10.4), TextVQA from 20.3 to 33.3
(+13.0), and DocVQA from 1.0 to 7.0 (+6.0, from a low baseline). This supports the hypothesis that complementary supervision signals (especially OCR
and grounding) add capabilities that single-task VQA supervision does not reliably induce.
This highlights the value of studying multi-task supervision effects in multimodal midtraining, emphasizing the importance of quantifying impact of diverse tasks on downstream performance.

\begin{figure*}[t]
    \centering
    \begin{minipage}[t]{0.48\textwidth}
        \centering
        \includegraphics[width=\linewidth]{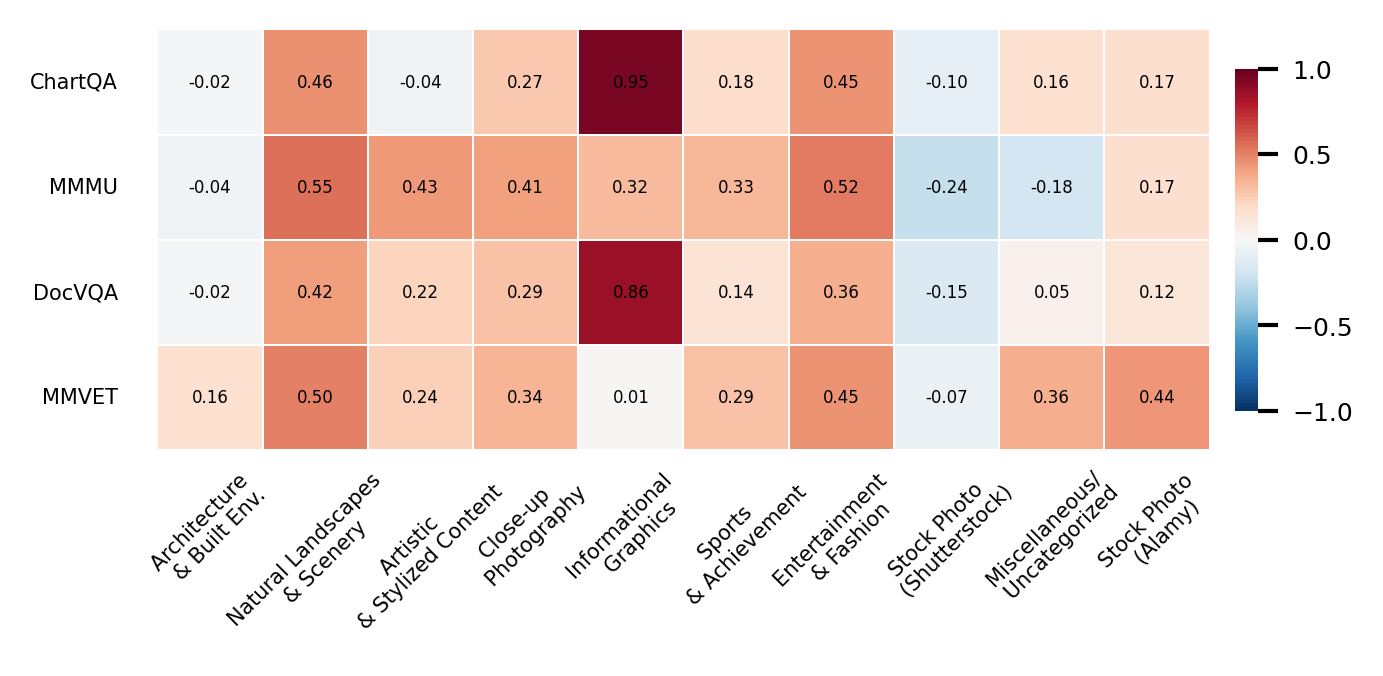}
        \caption{Image domain benchmark sensitivity. Heatmaps show per‑domain contribution scores for various benchmarks. Cell values indicate the relative contribution of each visual domain to benchmark performance (higher is better).}
        \label{fig:image_domain_sensitivity}
    \end{minipage}
    \hfill 
    \begin{minipage}[t]{0.48\textwidth}
        \centering
        \includegraphics[width=\linewidth]{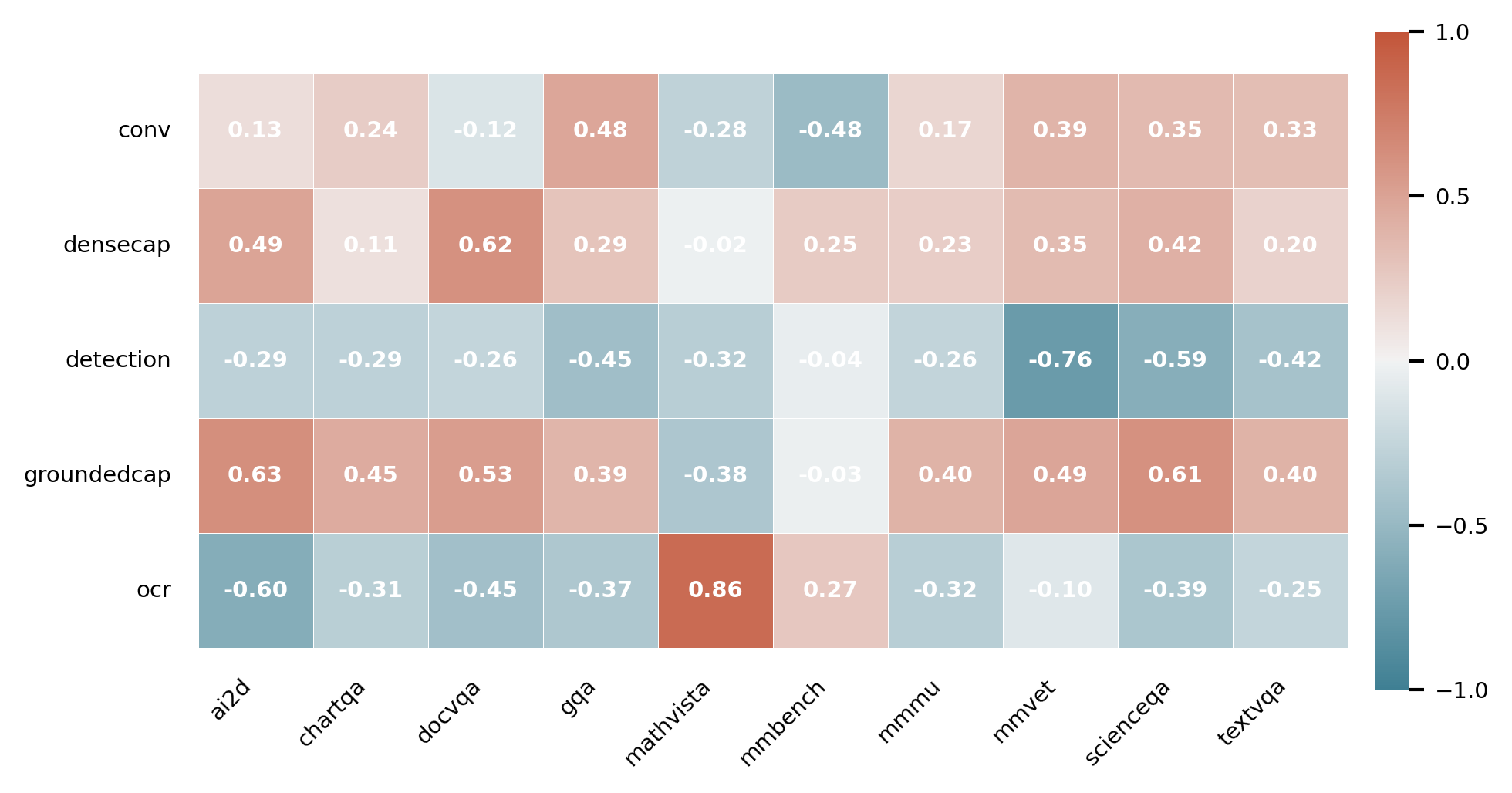}
        \caption{Task Domain–benchmark sensitivity. Heatmaps show per‑task domain contribution scores for 10 benchmarks. Cell values indicate the relative contribution of each task domain to benchmark performance (higher is better).}
        \label{fig:task_domain_sensitivity}
    \end{minipage}
\end{figure*}

\paragraph{Benchmark performance is highly domain- and task-sensitive.}
To understand \emph{why} different mixtures help different targets, we analyze benchmark sensitivity along two independent axes:
task type (supervision) and image concept. We compute the contribution score using the spearman correlation strength between domain weights and downstream performance. 
\begin{itemize}[noitemsep, topsep=0pt, leftmargin=10pt]
\item \textbf{Image-concept sensitivity.} As shown in Figure~\ref{fig:image_domain_sensitivity}, ChartQA and DocVQA are dominated by
\emph{Informational Graphics}, with very large positive contributions (0.95 and 0.86), consistent with their reliance on structured charts
and document-like layouts; other domains play secondary roles. In contrast, MMMU exhibits a broad, multi-domain dependence with strong positive
contributions spread across Natural Landscapes \& Scenery (0.55), Entertainment \& Fashion (0.52), Artistic \& Stylized Content (0.43),
and Close-up Photography (0.41), suggesting that it rewards visual diversity and broader world knowledge. MM-Vet shows a different profile:
it gains most from Natural Landscapes (0.50), Entertainment \& Fashion (0.45), and also Stock Photo (Alamy) (0.44) and
Miscellaneous/Uncategorized (0.36), while Informational Graphics contributes almost nothing (0.01). Finally, Stock Photo (Shutterstock)
is consistently negative across benchmarks (down to $-0.24$ on MMMU), indicating that some stock-photo distributions are weakly aligned
(or mildly harmful) for these evaluation targets.

\item \textbf{Task-type sensitivity.} Figure~\ref{fig:task_domain_sensitivity} shows that supervision choices also induce sharply different outcomes.
OCR supervision is highly specialized: it strongly benefits MathVista (0.86) and mildly helps MMBench (0.27), but is negative for most other
benchmarks. Grounded captioning is broadly beneficial: AI2D, ChartQA, MMMU, MM-Vet, ScienceQA, and TextVQA all peak with grounded captions
(0.63, 0.45, 0.40, 0.49, 0.61, 0.40). Dense captioning is the strongest signal for DocVQA (0.62, higher than grounded captioning at 0.53),
and conversational supervision is most helpful for GQA (0.48). Detection supervision is consistently harmful (negative for every benchmark,
with the largest drop on MM-Vet: $-0.76$). Overall, these maps show that \emph{there is no single best mixture}: chart/document-centric
benchmarks reward specialization, while broad-coverage evaluation often rewards diversity. 
\end{itemize}

\begin{figure}[t]
    \centering
    \begin{minipage}[b]{0.48\linewidth}
        \centering
        \includegraphics[width=\linewidth]{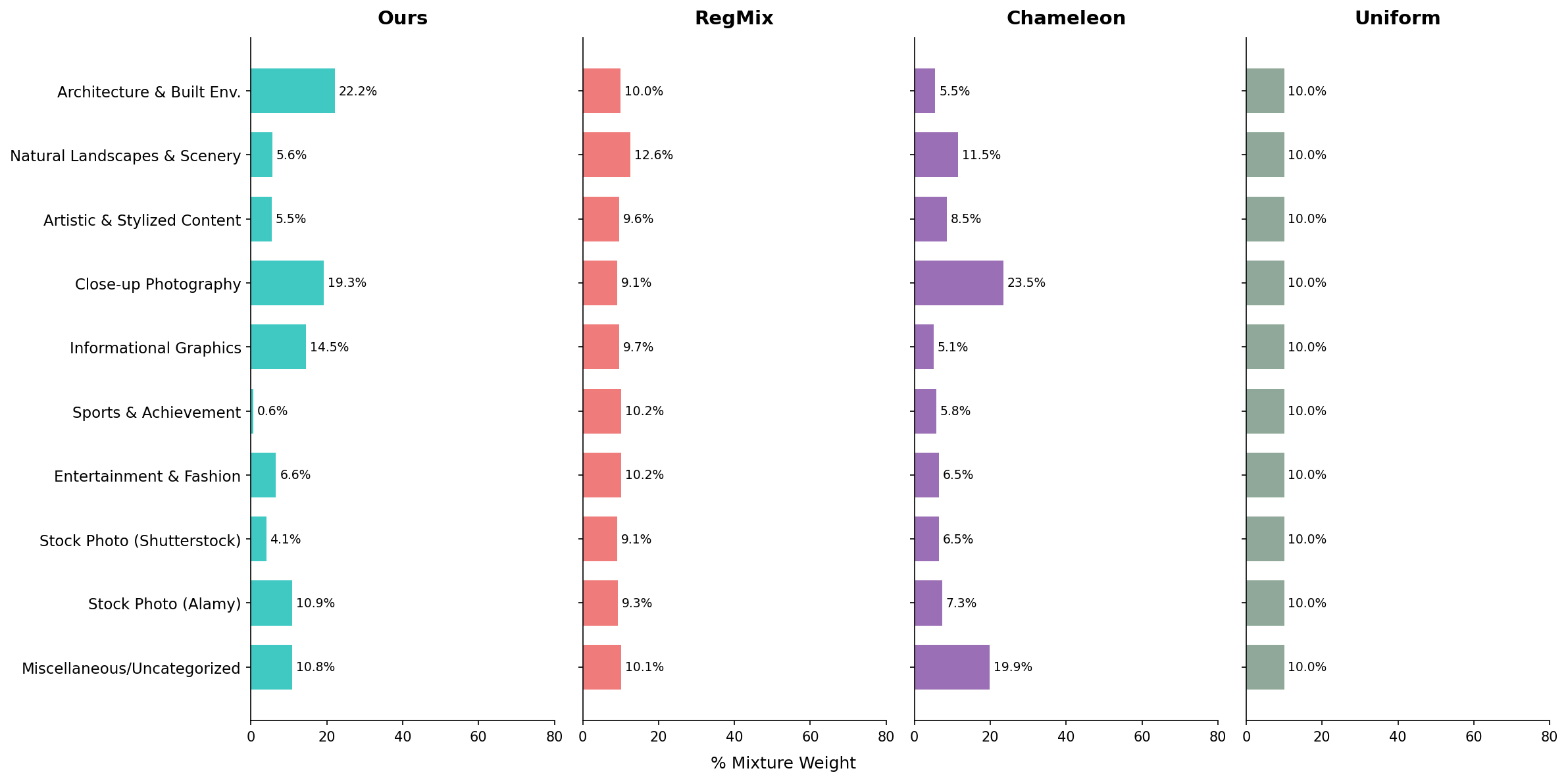}
        \centerline{(a)}
    \end{minipage}
    \hfill
    \begin{minipage}[b]{0.48\linewidth}
        \centering
        \includegraphics[width=\linewidth]{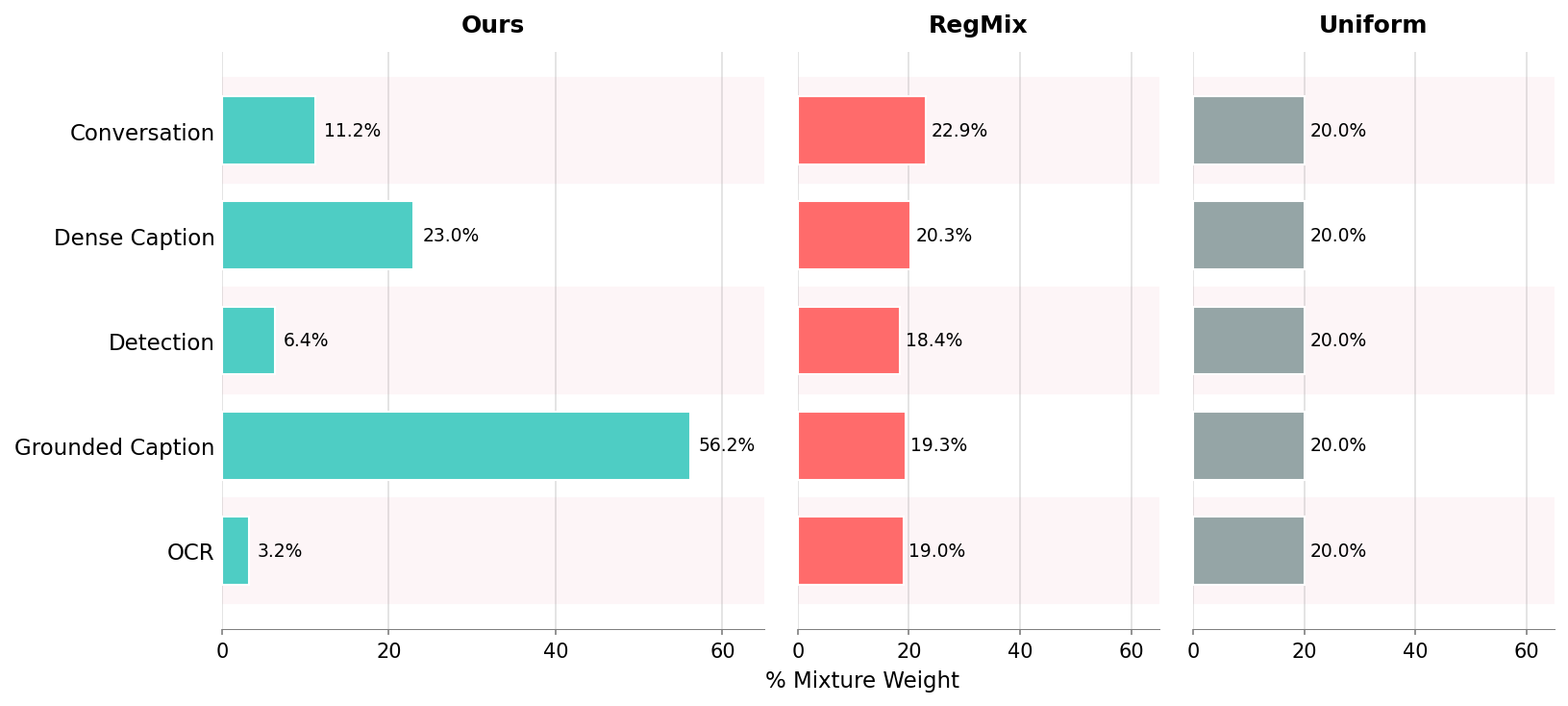}
        \centerline{(b)}
    \end{minipage}
    \caption{(a) Image concept domain weight distribution comparison across data mixing strategies. (b) Task type weight distribution comparison across different data mixing strategies. Our method assigns higher weight to Grounded Caption (56.2\%) compared to RegMix and Uniform baselines, which maintain near-uniform distributions across tasks.}
    \label{fig:optimal_weight_task_image}
\end{figure}

\paragraph{Optimal mixtures analysis} MixAtlas learns a sparse, interpretable recipe rather than a near-uniform one shown in Figure~\ref{fig:optimal_weight_task_image}. On the image-concept axis, it emphasizes Architecture \& Built Environment (22.2\%), Close-up Photography (19.3\%), and Informational Graphics (14.5\%), which matches Figure~\ref{fig:image_domain_sensitivity}: Informational Graphics is the strongest positive domain for ChartQA (0.95) and DocVQA (0.86). This specialization explains the large concept-axis gains in Table~\ref{tab:qwen2_7b_task_vs_image_concept} on DocVQA (15.8$\rightarrow$46.9), TextVQA (30.0$\rightarrow$50.9), and ChartQA (38.2$\rightarrow$41.8), yielding the best average of 47.18 (+17.6\%). On the task axis, MixAtlas puts most weight on Grounded Caption (56.2\%) and Dense Caption (23.0\%), consistent with Figure~\ref{fig:task_domain_sensitivity}, where grounded captioning is broadly helpful and dense captioning is strongest for DocVQA; accordingly, the task recipe reaches the best Qwen2-7B average of 50.68 (+8.5\%) and also transfers to Qwen2.5-7B with gains on GQA, DocVQA, and TextVQA.

\paragraph{Benchmark trade-off discussion} Figure~\ref{fig:image_domain_sensitivity} and Figure~\ref{fig:task_domain_sensitivity} also explain why MixAtlas does not improve every benchmark simultaneously. Figure~\ref{fig:image_domain_sensitivity} shows that chart/document benchmarks reward specialized concepts, whereas broader benchmarks such as MMMU benefit from more diverse visual domains; Figure~\ref{fig:task_domain_sensitivity} similarly shows benchmark-specific supervision needs, with grounded captioning helping ScienceQA, OCR helping MathVista, and conversational supervision helping GQA. As a result, recipes that favor document/text-oriented concepts and grounded or dense captions deliver large gains on DocVQA, TextVQA, and ChartQA, but can hurt broader benchmarks such as MMBench-EN or MathVista, as seen in Table~\ref{tab:qwen2_7b_task_vs_image_concept} and Table~\ref{tab:qwen2_5_7b_task_vs_image_concept}. This becomes even clearer on Qwen2.5-7B, where concept reweighting still helps document tasks but yields only a modest +1.0\% average gain because several broader benchmarks decline. These drops are therefore the expected cost of target-specific optimization, and MixAtlas makes that trade-off controllable by allowing users to reweight objectives or impose benchmark constraints during search.

%% file: sections/5_conclusion.tex
\section{Conclusion} \label{section:conclusion}
We presented \framework, an uncertainty-aware framework for optimizing multimodal midtraining data mixtures in a way that is both \emph{interpretable} and \emph{compute-efficient}.
MixAtlas converts heterogeneous midtraining corpora into a controllable data decomposition with two axis---task supervision and image concept---and uses proxy training with a probabilistic surrogate to explore the mixture space under a limited budget. We demonstrate that recipes discovered on small proxy models can transfer effectively to 7B-scale training, making mixture optimization practical in real-world compute regimes. Across diverse benchmarks, the resulting recipes improve average benchmark performance over existing baselines and increase training efficiency, reaching the same training loss in far fewer optimization steps. Our sensitivity analysis reveals that no single mixture dominates across all benchmarks, suggesting that future MLLMs training pipelines may benefit from maintaining a library of specialized recipes selected per-deployment target, rather than searching for a single universal mixture.

\section*{Limitations} 

While MixAtlas provides an interpretable and compute-efficient approach to multimodal mixture optimization, it has several limitations.

\textbf{Disjoint optimization across axes.} In this work, we treat the two decomposition axes---task type and image concept---as parallel ``knobs'' and optimize mixtures on each axis separately.
We do \emph{not} optimize the full joint cross-product portfolio (task $\times$ concept), which could capture important interaction effects (e.g., OCR on document-like images versus OCR on natural photos).
Joint optimization is attractive but increases the dimensionality of the search space; developing scalable joint methods (e.g., hierarchical/low-rank parameterizations or structured priors for the surrogate) is an important next step.

\textbf{Limited model and pipeline coverage.} Our experiments focus on LLaVA-Next style midtraining with Qwen-family backbones (Qwen2-7B and Qwen2.5-7B), and proxy runs based on smaller Qwen variants.
Although we demonstrate transfer across scales and across language models (Qwen2 $\rightarrow$ Qwen2.5), it remains to be seen how well the discovered recipes generalize to other language backbones (e.g., LLaMA-family), other vision encoders, and other MLLM training pipelines.
This is by design, since we want to discover recipes for a particular model, and believe that MixAtlas allows users to curate benchmark specific optimal recipes, however, discovering mixtures that transfer well across all different model scales and types is something we leave for future work.

%% file: main.bbl
\begin{thebibliography}{38}
\providecommand{\natexlab}[1]{#1}
\providecommand{\url}[1]{\texttt{#1}}
\expandafter\ifx\csname urlstyle\endcsname\relax
  \providecommand{\doi}[1]{doi: #1}\else
  \providecommand{\doi}{doi: \begingroup \urlstyle{rm}\Url}\fi

\bibitem[Ainslie et~al.(2023)Ainslie, Lee-Thorp, De~Jong, Zemlyanskiy, Lebr{\'o}n, and Sanghai]{ainslie2023gqa}
Joshua Ainslie, James Lee-Thorp, Michiel De~Jong, Yury Zemlyanskiy, Federico Lebr{\'o}n, and Sumit Sanghai.
\newblock Gqa: Training generalized multi-query transformer models from multi-head checkpoints.
\newblock \emph{arXiv preprint arXiv:2305.13245}, 2023.

\bibitem[Albalak et~al.(2023)Albalak, Pan, Raffel, and Wang]{wang2023datamixture}
Alon Albalak, Liangming Pan, Colin Raffel, and William~Yang Wang.
\newblock Efficient online data mixing for language model pre-training.
\newblock \emph{arXiv preprint arXiv:2312.02406}, 2023.
\newblock URL \url{https://arxiv.org/abs/2312.02406}.

\bibitem[Bai et~al.(2024)Bai, Liang, Wan, Yang, Li, Wang, Cui, He, Yuan, and Zhang]{Baietal2024}
Tianyi Bai, Hao Liang, Binwang Wan, Ling Yang, Bozhou Li, Yifan Wang, Bin Cui, Conghui He, Binhang Yuan, and Wentao Zhang.
\newblock A survey of multimodal large language model from a data-centric perspective.
\newblock \emph{arXiv.org}, 2024.

\bibitem[Beyer et~al.(2024)Beyer, Steiner, Pinto, Kolesnikov, Wang, Salz, Neumann, Alabdulmohsin, Tschannen, Bugliarello, et~al.]{beyer2024paligemma}
Lucas Beyer, Andreas Steiner, Andr{\'e}~Susano Pinto, Alexander Kolesnikov, Xiao Wang, Daniel Salz, Maxim Neumann, Ibrahim Alabdulmohsin, Michael Tschannen, Emanuele Bugliarello, et~al.
\newblock Paligemma: A versatile 3b vlm for transfer.
\newblock \emph{arXiv preprint arXiv:2407.07726}, 2024.

\bibitem[Changpinyo et~al.(2021)Changpinyo, Sharma, Ding, and Soricut]{changpinyo2021cc12m}
Soravit Changpinyo, Piyush Sharma, Nan Ding, and Radu Soricut.
\newblock {Conceptual 12M}: Pushing web-scale image-text pre-training to recognize long-tail visual concepts.
\newblock In \emph{CVPR}, 2021.

\bibitem[Chen et~al.(2023)Chen, Li, Dong, Zhang, He, Wang, Zhao, and Lin]{chen2023sharegpt4v}
Lin Chen, Jinsong Li, Xiaoyi Dong, Pan Zhang, Conghui He, Jiaqi Wang, Feng Zhao, and Dahua Lin.
\newblock Sharegpt4v: Improving large multi-modal models with better captions.
\newblock \emph{arXiv preprint arXiv:2311.12793}, 2023.
\newblock URL \url{https://arxiv.org/abs/2311.12793}.

\bibitem[Deitke et~al.(2025)Deitke, Clark, Lee, Tripathi, Yang, Park, Salehi, Muennighoff, Lo, Soldaini, et~al.]{deitke2025molmo}
Matt Deitke, Christopher Clark, Sangho Lee, Rohun Tripathi, Yue Yang, Jae~Sung Park, Mohammadreza Salehi, Niklas Muennighoff, Kyle Lo, Luca Soldaini, et~al.
\newblock Molmo and pixmo: Open weights and open data for state-of-the-art vision-language models.
\newblock In \emph{Proceedings of the Computer Vision and Pattern Recognition Conference}, pages 91--104, 2025.

\bibitem[Diao et~al.(2025)Diao, Yang, Fu, Dong, SU, Kliegl, CHEN, Belcak, Suhara, Yin, Patwary, Lin, Kautz, and Molchanov]{diao2025nemotronclimb}
Shizhe Diao, Yu~Yang, Yonggan Fu, Xin Dong, Dan SU, Markus Kliegl, ZIJIA CHEN, Peter Belcak, Yoshi Suhara, Hongxu Yin, Mostofa Patwary, Yingyan~Celine Lin, Jan Kautz, and Pavlo Molchanov.
\newblock Nemotron-{CLIMB}: Clustering-based iterative data mixture bootstrapping for language model pre-training.
\newblock In \emph{The Thirty-ninth Annual Conference on Neural Information Processing Systems Datasets and Benchmarks Track}, 2025.
\newblock URL \url{https://openreview.net/forum?id=aBlqKPkc4a}.

\bibitem[Du et~al.(2022)Du, Huang, Dai, et~al.]{du2022mixturejoint}
Nan Du, Yanping Huang, Andrew~M. Dai, et~al.
\newblock Glam: Efficient scaling of language models with mixture-of-experts.
\newblock In \emph{Proceedings of the 39th International Conference on Machine Learning}, volume 162 of \emph{Proceedings of Machine Learning Research}, pages 5547--5569. PMLR, 2022.
\newblock URL \url{https://proceedings.mlr.press/v162/du22c.html}.

\bibitem[Gadre et~al.(2023)Gadre, Ilharco, Fang, Hayase, Smyrnis, Nguyen, Marten, Wortsman, Ghosh, Zhang, and et~al.]{gadre2023datacomp}
Samir~Yitzhak Gadre, Gabriel Ilharco, Alex Fang, Jonathan Hayase, Georgios Smyrnis, Thao Nguyen, Ryan Marten, Mitchell Wortsman, Dhruba Ghosh, Jieyu Zhang, and et~al.
\newblock Datacomp: In search of the next generation of multimodal datasets.
\newblock \emph{arXiv preprint arXiv:2304.14108}, 2023.
\newblock URL \url{https://arxiv.org/abs/2304.14108}.

\bibitem[Kang et~al.(2024)Kang, Sun, Wen, Chen, Song, Mahmood, and Jia]{kang2024autoscale}
Feiyang Kang, Yifan Sun, Bingbing Wen, Si~Chen, Dawn Song, Rafid Mahmood, and Ruoxi Jia.
\newblock Autoscale: Scale-aware data mixing for pre-training llms.
\newblock \emph{arXiv preprint arXiv:2407.20177}, 2024.

\bibitem[Kembhavi et~al.(2016)Kembhavi, Salvato, Kolve, Seo, Hajishirzi, and Farhadi]{Kembhavi2016ADI}
Aniruddha Kembhavi, Michael Salvato, Eric Kolve, Minjoon Seo, Hannaneh Hajishirzi, and Ali Farhadi.
\newblock A diagram is worth a dozen images.
\newblock \emph{ArXiv}, abs/1603.07396, 2016.
\newblock URL \url{https://api.semanticscholar.org/CorpusID:2682274}.

\bibitem[Liu et~al.(2025)Liu, Neubig, and Xiong]{Liu2025MidtrainingBPAU}
Emmy Liu, Graham Neubig, and Chenyan Xiong.
\newblock Midtraining bridges pretraining and posttraining distributions.
\newblock \emph{ArXiv}, abs/2510.14865, 2025.
\newblock URL \url{https://api.semanticscholar.org/CorpusId:282138804}.

\bibitem[Liu et~al.(2023)Liu, Li, Wu, and Lee]{liu2023llava}
Haotian Liu, Chunyuan Li, Qingyang Wu, and Yong~Jae Lee.
\newblock Visual instruction tuning.
\newblock \emph{arXiv preprint arXiv:2304.08485}, 2023.
\newblock URL \url{https://arxiv.org/abs/2304.08485}.
\newblock NeurIPS 2023 (Oral).

\bibitem[Liu et~al.(2024{\natexlab{a}})Liu, Zheng, Muennighoff, Zeng, Dou, Pang, Jiang, and Lin]{liu2024regmix}
Qian Liu, Xiaosen Zheng, Niklas Muennighoff, Guangtao Zeng, Longxu Dou, Tianyu Pang, Jing Jiang, and Min Lin.
\newblock {RegMix}: Data mixture as regression for language model pre-training.
\newblock \emph{arXiv preprint arXiv:2407.01492}, 2024{\natexlab{a}}.
\newblock URL \url{https://arxiv.org/abs/2407.01492}.
\newblock ICLR 2025 (to appear).

\bibitem[Liu et~al.(2024{\natexlab{b}})Liu, Duan, Zhang, Li, Zhang, Zhao, Yuan, Wang, He, Liu, et~al.]{liu2024mmbench}
Yuan Liu, Haodong Duan, Yuanhan Zhang, Bo~Li, Songyang Zhang, Wangbo Zhao, Yike Yuan, Jiaqi Wang, Conghui He, Ziwei Liu, et~al.
\newblock Mmbench: Is your multi-modal model an all-around player?
\newblock In \emph{European conference on computer vision}, pages 216--233. Springer, 2024{\natexlab{b}}.

\bibitem[Lu et~al.(2022)Lu, Mishra, Xia, Qiu, Chang, Zhu, Tafjord, Clark, and Kalyan]{lu2022learn}
Pan Lu, Swaroop Mishra, Tony Xia, Liang Qiu, Kai-Wei Chang, Song-Chun Zhu, Oyvind Tafjord, Peter Clark, and Ashwin Kalyan.
\newblock Learn to explain: Multimodal reasoning via thought chains for science question answering.
\newblock In \emph{The 36th Conference on Neural Information Processing Systems (NeurIPS)}, 2022.

\bibitem[Lu et~al.(2024)Lu, Bansal, Xia, Liu, Li, Hajishirzi, Cheng, Chang, Galley, and Gao]{lu2024mathvista}
Pan Lu, Hritik Bansal, Tony Xia, Jiacheng Liu, Chunyuan Li, Hannaneh Hajishirzi, Hao Cheng, Kai-Wei Chang, Michel Galley, and Jianfeng Gao.
\newblock Mathvista: Evaluating mathematical reasoning of foundation models in visual contexts.
\newblock In \emph{International Conference on Learning Representations (ICLR)}, 2024.

\bibitem[Masry et~al.(2022)Masry, Long, Tan, Joty, and Hoque]{masry-etal-2022-chartqa}
Ahmed Masry, Do~Long, Jia~Qing Tan, Shafiq Joty, and Enamul Hoque.
\newblock {C}hart{QA}: A benchmark for question answering about charts with visual and logical reasoning.
\newblock In \emph{Findings of the Association for Computational Linguistics: ACL 2022}, pages 2263--2279, Dublin, Ireland, May 2022. Association for Computational Linguistics.
\newblock \doi{10.18653/v1/2022.findings-acl.177}.
\newblock URL \url{https://aclanthology.org/2022.findings-acl.177}.

\bibitem[Mathew et~al.(2021)Mathew, Karatzas, and Jawahar]{mathew2021docvqa}
Minesh Mathew, Dimosthenis Karatzas, and CV~Jawahar.
\newblock Docvqa: A dataset for vqa on document images.
\newblock In \emph{Proceedings of the IEEE/CVF winter conference on applications of computer vision}, pages 2200--2209, 2021.

\bibitem[McKinzie et~al.(2024)McKinzie, Gan, Fauconnier, Dodge, Zhang, Dufter, Shah, Du, Peng, Weers, Belyi, Zhang, Singh, Kang, Jain, H{\`e}, Schwarzer, Gunter, Kong, Zhang, Wang, Wang, Du, Lei, Wiseman, Yin, Lee, Wang, Pang, Grasch, Toshev, and Yang]{McKinzie2024MM1MAC}
Brandon McKinzie, Zhe Gan, J.~Fauconnier, Sam Dodge, Bowen Zhang, Philipp Dufter, Dhruti Shah, Xianzhi Du, Futang Peng, Floris Weers, Anton Belyi, Haotian Zhang, Karanjeet Singh, Doug Kang, Ankur Jain, Hongyu H{\`e}, Max Schwarzer, Tom Gunter, Xiang Kong, Aonan Zhang, Jianyu Wang, Chong Wang, Nan Du, Tao Lei, Sam Wiseman, Guoli Yin, Mark Lee, Zirui Wang, Ruoming Pang, Peter Grasch, Alexander Toshev, and Yinfei Yang.
\newblock Mm1: Methods, analysis \& insights from multimodal llm pre-training.
\newblock \emph{ArXiv}, abs/2403.09611, 2024.
\newblock URL \url{https://api.semanticscholar.org/CorpusId:268384865}.

\bibitem[Peng et~al.(2019)Peng, Bai, Xia, Huang, Saenko, and Wang]{peng2019domain}
Xingchao Peng, Qinxun Bai, Xide Xia, Zijun Huang, Kate Saenko, and Bo~Wang.
\newblock Moment matching for multi-source domain adaptation.
\newblock In \emph{Proceedings of the IEEE/CVF International Conference on Computer Vision (ICCV)}, 2019.
\newblock URL \url{https://openaccess.thecvf.com/content_ICCV_2019/html/Peng_Moment_Matching_for_Multi-Source_Domain_Adaptation_ICCV_2019_paper.html}.

\bibitem[Radford et~al.(2021)Radford, Kim, Hallacy, Ramesh, Goh, Agarwal, Sastry, Askell, Mishkin, Clark, et~al.]{radford2021learning}
Alec Radford, Jong~Wook Kim, Chris Hallacy, Aditya Ramesh, Gabriel Goh, Sandhini Agarwal, Girish Sastry, Amanda Askell, Pamela Mishkin, Jack Clark, et~al.
\newblock Learning transferable visual models from natural language supervision.
\newblock In \emph{International conference on machine learning}, pages 8748--8763. PmLR, 2021.

\bibitem[Roth et~al.(2024)Roth, Udandarao, Dziadzio, Prabhu, Cherti, Vinyals, Henaff, Albanie, Bethge, and Akata]{roth2024a}
Karsten Roth, Vishaal Udandarao, Sebastian Dziadzio, Ameya Prabhu, Mehdi Cherti, Oriol Vinyals, Olivier~J Henaff, Samuel Albanie, Matthias Bethge, and Zeynep Akata.
\newblock A practitioner's guide to continual multimodal pretraining.
\newblock In \emph{NeurIPS 2024 Workshop on Scalable Continual Learning for Lifelong Foundation Models}, 2024.
\newblock URL \url{https://openreview.net/forum?id=gkyosluSbR}.

\bibitem[Sharma et~al.(2018)Sharma, Ding, Goodman, and Soricut]{sharma2018conceptual}
Piyush Sharma, Nan Ding, Sebastian Goodman, and Radu Soricut.
\newblock Conceptual captions: A cleaned, hypernymed, image alt-text dataset for automatic image captioning.
\newblock In \emph{Proceedings of ACL}, 2018.

\bibitem[Shukor et~al.(2025)Shukor, Bethune, Busbridge, Grangier, Fini, El-Nouby, and Ablin]{shukor2025scaling}
Mustafa Shukor, Louis Bethune, Dan Busbridge, David Grangier, Enrico Fini, Alaaeldin El-Nouby, and Pierre Ablin.
\newblock Scaling laws for optimal data mixtures.
\newblock \emph{arXiv preprint arXiv:2507.09404}, 2025.

\bibitem[Singh et~al.(2019)Singh, Natarjan, Shah, Jiang, Chen, Parikh, and Rohrbach]{singh2019towards}
Amanpreet Singh, Vivek Natarjan, Meet Shah, Yu~Jiang, Xinlei Chen, Devi Parikh, and Marcus Rohrbach.
\newblock Towards vqa models that can read.
\newblock In \emph{Proceedings of the IEEE Conference on Computer Vision and Pattern Recognition}, pages 8317--8326, 2019.

\bibitem[Snoek et~al.(2012)Snoek, Larochelle, and Adams]{snoek2012practical}
Jasper Snoek, Hugo Larochelle, and Ryan~P. Adams.
\newblock Practical bayesian optimization of machine learning algorithms.
\newblock In \emph{Advances in Neural Information Processing Systems}, volume~25, pages 2951--2959, 2012.
\newblock URL \url{https://proceedings.neurips.cc/paper/2012/file/05311655a15b75fab86956663e1819cd-Paper.pdf}.

\bibitem[Wen et~al.(2023)Wen, Yang, Wang, Gan, Howe, and Wang]{wen2023infovisdial}
Bingbing Wen, Zhengyuan Yang, Jianfeng Wang, Zhe Gan, Bill Howe, and Lijuan Wang.
\newblock Infovisdial: An informative visual dialogue dataset by bridging large multimodal and language models.
\newblock \emph{arXiv preprint arXiv:2312.13503}, 2023.

\bibitem[Wettig et~al.(2025)Wettig, Lo, Min, Hajishirzi, Chen, and Soldaini]{wettig2025organize}
Alexander Wettig, Kyle Lo, Sewon Min, Hannaneh Hajishirzi, Danqi Chen, and Luca Soldaini.
\newblock Organize the web: Constructing domains enhances pre-training data curation.
\newblock In \emph{Forty-second International Conference on Machine Learning}, 2025.
\newblock URL \url{https://openreview.net/forum?id=boSqwdvJVC}.

\bibitem[Xie et~al.(2023)Xie, Pham, Dong, Du, Liu, Lu, Liang, Le, Ma, and Yu]{Xie2023DoReMi}
Sang~Michael Xie, Hieu Pham, Xuanyi Dong, Nan Du, Hanxiao Liu, Yifeng Lu, Percy Liang, Quoc~V. Le, Tengyu Ma, and Adams~Wei Yu.
\newblock {DoReMi}: Optimizing data mixtures speeds up language model pretraining.
\newblock In \emph{Advances in Neural Information Processing Systems}, 2023.
\newblock URL \url{https://proceedings.neurips.cc/paper_files/paper/2023/file/dcba6be91359358c2355cd920da3fcbd-Paper-Conference.pdf}.

\bibitem[Xie et~al.(2025)Xie, Tonin, and Cevher]{xie2025chameleon}
Wanyun Xie, Francesco Tonin, and Volkan Cevher.
\newblock Chameleon: A flexible data-mixing framework for language model pretraining and finetuning.
\newblock In \emph{Forty-second International Conference on Machine Learning}, 2025.
\newblock URL \url{https://openreview.net/forum?id=mDxarRaTY9}.

\bibitem[Yang et~al.(2025{\natexlab{a}})Yang, Li, Yang, Zhang, Hui, Zheng, Yu, Gao, Huang, Lv, et~al.]{yang2025qwen3}
An~Yang, Anfeng Li, Baosong Yang, Beichen Zhang, Binyuan Hui, Bo~Zheng, Bowen Yu, Chang Gao, Chengen Huang, Chenxu Lv, et~al.
\newblock Qwen3 technical report.
\newblock \emph{arXiv preprint arXiv:2505.09388}, 2025{\natexlab{a}}.

\bibitem[Yang et~al.(2022)Yang, Hu, Babuschkin, Sidor, Liu, Farhi, Ryder, Pachocki, Chen, and Gao]{yang2022mutransfer}
Greg Yang, Edward~J. Hu, Igor Babuschkin, Szymon Sidor, Xiaodong Liu, David Farhi, Nick Ryder, Jakub Pachocki, Weizhu Chen, and Jianfeng Gao.
\newblock Tensor programs v: Tuning large neural networks via zero-shot hyperparameter transfer.
\newblock \emph{arXiv preprint arXiv:2203.03466}, 2022.
\newblock URL \url{https://arxiv.org/abs/2203.03466}.

\bibitem[Yang et~al.(2025{\natexlab{b}})Yang, Lee, Feng, Zhao, Wen, Liu, Tsvetkov, and Howe]{yang2025escaping}
Yiwei Yang, Chung~Peng Lee, Shangbin Feng, Dora Zhao, Bingbing Wen, Anthony~Zhe Liu, Yulia Tsvetkov, and Bill Howe.
\newblock Escaping the spuriverse: Can large vision-language models generalize beyond seen spurious correlations?
\newblock In \emph{The Thirty-ninth Annual Conference on Neural Information Processing Systems Datasets and Benchmarks Track}, 2025{\natexlab{b}}.
\newblock URL \url{https://openreview.net/forum?id=es2NkPKFCB}.

\bibitem[Yao et~al.(2025)Yao, Hu, Yi, Han, Feng, Yang, Wen, Krishna, Wang, Tsvetkov, et~al.]{yao2025mmmg}
Jihan Yao, Yushi Hu, Yujie Yi, Bin Han, Shangbin Feng, Guang Yang, Bingbing Wen, Ranjay Krishna, Lucy~Lu Wang, Yulia Tsvetkov, et~al.
\newblock Mmmg: a comprehensive and reliable evaluation suite for multitask multimodal generation.
\newblock \emph{arXiv preprint arXiv:2505.17613}, 2025.

\bibitem[Yu et~al.(2024)Yu, Yang, Li, Wang, Lin, Liu, Wang, and Wang]{yu2024mm}
Weihao Yu, Zhengyuan Yang, Linjie Li, Jianfeng Wang, Kevin Lin, Zicheng Liu, Xinchao Wang, and Lijuan Wang.
\newblock Mm-vet: Evaluating large multimodal models for integrated capabilities.
\newblock In \emph{International conference on machine learning}. PMLR, 2024.

\bibitem[Yue et~al.(2024)Yue, Ni, Zhang, Zheng, Liu, Zhang, Stevens, Jiang, Ren, Sun, Wei, Yu, Yuan, Sun, Yin, Zheng, Yang, Liu, Huang, Sun, Su, and Chen]{yue2023mmmu}
Xiang Yue, Yuansheng Ni, Kai Zhang, Tianyu Zheng, Ruoqi Liu, Ge~Zhang, Samuel Stevens, Dongfu Jiang, Weiming Ren, Yuxuan Sun, Cong Wei, Botao Yu, Ruibin Yuan, Renliang Sun, Ming Yin, Boyuan Zheng, Zhenzhu Yang, Yibo Liu, Wenhao Huang, Huan Sun, Yu~Su, and Wenhu Chen.
\newblock Mmmu: A massive multi-discipline multimodal understanding and reasoning benchmark for expert agi.
\newblock In \emph{Proceedings of CVPR}, 2024.

\end{thebibliography}
